\documentclass[letter, 10pt]{IEEEtran}
\IEEEoverridecommandlockouts
\usepackage{cite}
\usepackage{amsmath,amssymb,amsfonts}
\usepackage{amsthm}
\usepackage[]{algorithm2e}
\usepackage{algpseudocode}
\usepackage{graphicx}
\usepackage{textcomp}
\usepackage{xcolor}
\usepackage{hyperref}
\usepackage[caption=false,font=footnotesize]{subfig}
\usepackage{flushend}

\newtheorem{defn}{Definition}[section]

\begin{document}

\title{SMART-3D: Three-Dimensional Self-Morphing Adaptive Replanning Tree\\
{\footnotesize \textsuperscript{}}\vspace{-18pt}
}

\author{\begin{tabular}{cccccccccc}
{Priyanshu Agrawal$^\dag$} & {Shalabh Gupta$^\dag$$^\star$} & {Zongyuan Shen$^\ddag$}\end{tabular}\vspace{-6pt}
\thanks {$^\dag$ Dept. of Electrical and Computer Engineering, University of Connecticut, Storrs, CT, USA.}
\thanks {$^\ddag$ Robotics Institute, Carnegie Mellon University, Pittsburgh, PA, USA.}
\thanks {$^\star$Corresponding author (email: shalabh.gupta@uconn.edu)} \vspace{-0pt}
\thanks {$^{\#}$Image generated by ChatGPT} \vspace{-3pt}
}

\maketitle

\thispagestyle{empty}

\begin{abstract}
This paper presents SMART-3D, an extension of the SMART algorithm to 3D environments. SMART-3D is a tree-based adaptive replanning algorithm for dynamic environments with fast moving obstacles. SMART-3D morphs the underlying tree to find a new path in real-time whenever the current path is blocked by obstacles. SMART-3D removed the grid decomposition requirement of the SMART algorithm by replacing the concept of hot-spots with that of hot-nodes, thus making it computationally efficient and scalable to 3D environments. The hot-nodes are nodes which allow for efficient reconnections to morph the existing tree to find a new safe and reliable path. The performance of SMART-3D is evaluated by extensive simulations in 2D and 3D environments populated with randomly moving dynamic obstacles. The results show that SMART-3D achieves high success rates and low replanning times, thus highlighting its suitability for real-time onboard applications.
\end{abstract}

\section{Introduction}

\begin{figure}
    \centering
    \includegraphics[width=0.90\linewidth]{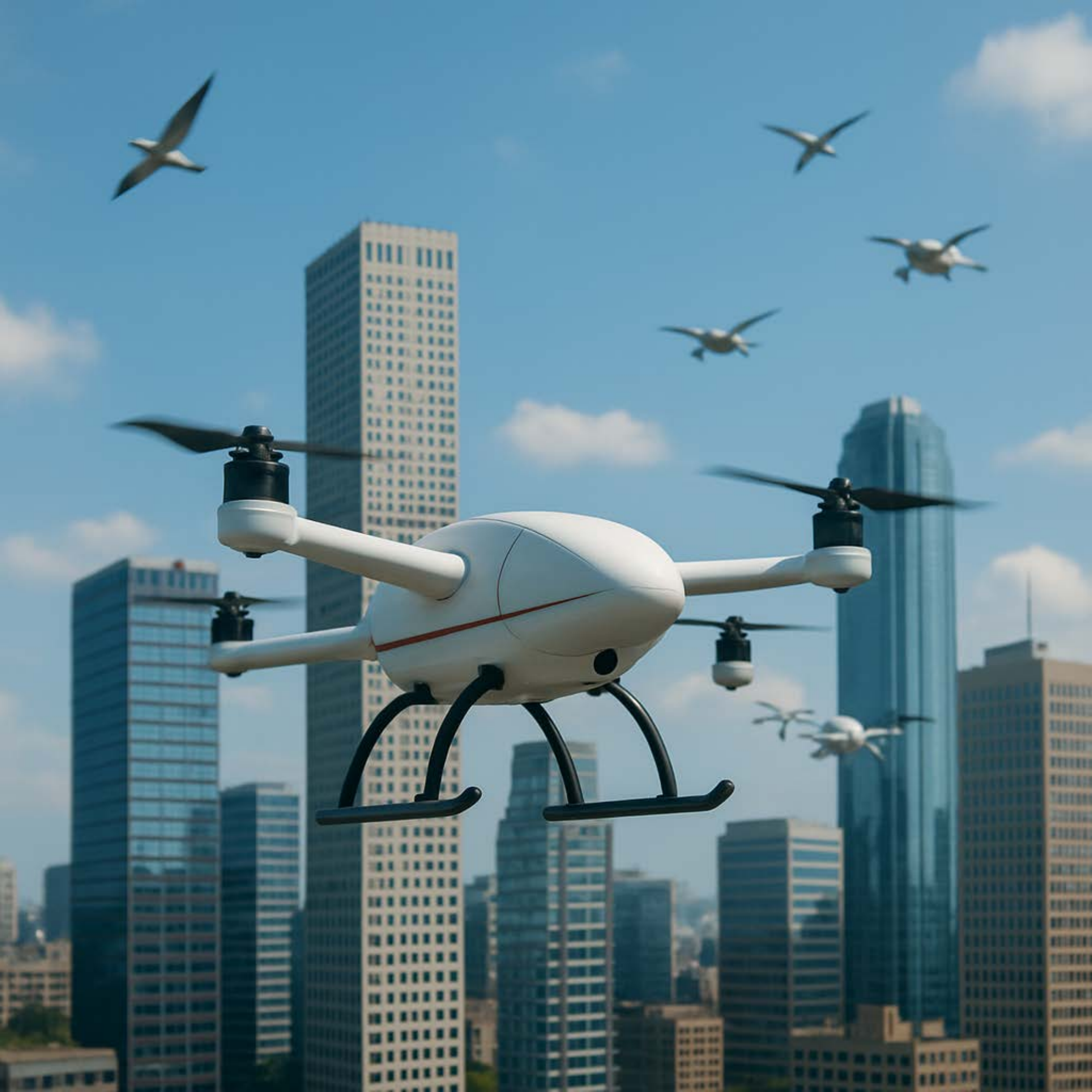}
    \caption{UAV flying in a city with both static and dynamic obstacles$^{\#}$.}
    \label{fig:UAVcity}\vspace{-10pt}
\end{figure}

Recent decades have seen significant growth of autonomous robots in supporting a diverse range of human operations. For example,  unmanned underwater vehicles (UUVs) are used in different applications such as marine life exploration~\cite{Katzschmann2018}\cite{Yuh2011_Applications_marinerobot}, underwater terrain mapping~\cite{shen2022ct,CStar25,palomeras2018,song2018,Negahdaripour2003}), oceanic data measurement (e.g., salinity) ~\cite{Somers2016}, biological sampling~\cite{Galloway2016_Softroboticgrippers}, oil spill cleaning~\cite{SGH13,Kumar2020}, object recognition~\cite{Foresti2002}, diver action recognition~\cite{YangWilson2023}, exploring mineral resources~\cite{Wakita2010_AUVResources}, target tracking~\cite{Shojaei2017, HGW2019, Hare2020},  mine hunting~\cite{Acar2003, mukherjee2011symbolic}, energy-constrained sorties~\cite{shen2020} and structural inspection~\cite{Foresti2001}. Similarly, unmanned air vehicles (UAVs) are used in different applications such as environmental monitoring~\cite{dunbabin2012robots}, surveillance~\cite{MGRW11}, terrain mapping~\cite{Fabricio24}, tree inspection~\cite{agriculture13020354}, smart agriculture~\cite{Moradi22} and bridge monitoring~\cite{PANIGATI2025106101}.
Most applications of autonomous robots consist of complex dynamic environments characterized by static as well as dynamic obstacles, where safe and reliable navigation becomes challenging. Fig. ~\ref{fig:UAVcity} shows an illustrative example, where a UAV is navigating in a city environment with buildings and other dynamic obstacles. 

\subsection{Motivation}
In order for the autonomous robots to operate effectively in human-centric real-world environments, they need to be able to react and adapt their paths~\cite{GRP09} in real-time to accommodate the changes in the environment. The standard path planning algorithms work offline to find an obstacle-free path that minimizes a given cost function, such as travel time. However, in dynamic environments, the robots require the ability to continuously replan in real-time due to moving obstacles. In addition to low travel times, these robots need to achieve low replanning times and high success rates (i.e., low probability of collision)~\cite{smart} for safe and reliable navigation.

\subsection{Literature Review}
There are two types of strategies for planning in dynamic environments: active and reactive. Active planning strategies predict future obstacle trajectories to inform their plans. However, trajectory predictions specially for multiple obstacles could become computationally expensive and inaccurate~\cite{HGW2019}, thus leading to unreliable paths. As such, reactive strategies have become popular, which use only current information of the environment and react to changes as they occur.

Technical literature consists of several tree-based replanning methods considering dynamic obstacles. These methods utilize different tree pruning and repair strategies to adapt the tree to dynamic obstacles and generate a new path. Such methods include Extended RRT (ERRT) \cite{bruce2002real},  
Dynamic RRT (DRRT)\cite{ferguson2006replanning}, Multipartite RRT (MPRRT) \cite{zucker2007multipartite}, RRT$^\text{X}$ \cite{otte2016rrtx}, Horizon-based Lazy RRT* (HLRRT*) \cite{chen2019horizon}, Efficient Bias-goal Factor RRT (EBGRRT) \cite{yuan2020efficient} and Multi-objective Dynamic RRT* (MODRRT*)\cite{qi2020}. There also exist search-based methods, such as 
D* Lite\cite{koenig2002d} and Lifelong planning A*\cite{koenig2004lifelong}. Some methods also incorporated steady~\cite{mittal2020rapid} and spatio-temporally varying~\cite{MittalGupta2019} currents for adaptive planning.

A recently proposed algorithm, called Self-Morphing Adaptive Replanning Tree (SMART)~\cite{smart}, performs fast reactive replanning for mobile robots in dynamic obstacle environments. A key distinct feature of SMART from the above tree-based methods is that it restricts the path feasibility check and tree-pruning to the robot's local neighborhood. Furthermore, SMART preserves the maximal tree structure via saving the multiple disjoint subtrees created after pruning infeasible colliding nodes and edges. These subtrees are then incrementally repaired and merged to form a morphed tree, which can produce the new collision-free path. The repair is done at hot-spots which are grid cells whose neighbors contain nodes of multiple-disjoint subtrees, thus providing avenues for easily reconnecting disjoint subtrees and perform repair. In rapidly changing environments, the algorithm outperformed the above tree-based and search-based reactive replanning algorithms in terms of travel time, success rate and replanning time ~\cite{smart}.  
A variant of SMART, called SMART-OC~\cite{SMARTOC25}, adapted SMART to  perform real-time time-risk optimal~\cite{Songgupta2019} replanning in response to dynamic obstacles and currents.

\subsection{Contributions}
SMART was developed for 2-dimensional environments, such as unmanned ground vehicles (UGVs) (e.g., wheel-base robots transporting goods in a factory) or unmanned surface vehicles (USVs) (e.g., autonomous boats operating offshore on an ocean surface). In this regard, this paper develops an extended version of SMART, called SMART-3D, which allows its deployment to robots (e.g., UAVs and UUVs) that operate in 3-dimensional environments. To facilitate computationally efficient operation in 3-D environments, SMART-3D replaced the concept of hot-spots that rely on grid decomposition for searching, with an equivalent concept of hot-nodes.

\section{SMART-3D Algorithm}
This section first presents a brief summary of the SMART-3D algorithm and then describes the necessary details. The main steps of SMART-3D are aligned with the steps of SMART~\cite{smart}. Initially, SMART-3D constructs a planning tree using the RRT* algorithm considering only the static obstacles. This tree is then used to compute the initial path of the robot. While navigating, the robot constantly checks if its path is blocked by nearby dynamic obstacles. If the path is blocked, SMART-3D prunes the risky nodes in the local neighborhood of the robot. This breaks the RRT* tree and could result in the formation of multiple disjoint subtrees. 

After pruning, SMART-3D performs the tree-repair to form a morphed tree, which can produce a new collision-free path. For tree-repair, it searches around the robot for critical nodes, called hot-nodes, whose neighboring nodes belong to atleast one other disjoint subtree. Then, the hot-nodes are ranked according to a distance-based utility heuristic. Next, the hot-nodes are incrementally selected based on their utility and connected to their neighbors in different subtrees, thus merging the disjoint subtrees until a new path is found to the goal. This replanning process is repeated every time the robot's path is blocked by nearby dynamic obstacles until the goal is reached. SMART-3D is computationally very efficient because it performs path feasibility checking, tree-pruning, and hot-spot searching in the local vicinity of the robot and it utilizes the maximum disjoint tree-structure after pruning for quick tree-repair, thus producing a new path in real-time. 

As compared to SMART~\cite{smart}, a key difference in SMART-3D is the removal of the grid decomposition of the space. SMART uses the grid decomposition to define and search for hot-spots. However, a grid-based search could become computationally expensive in 3D and higher dimensional spaces. As such, SMART-3D replaces the hot-spots with a functionally equivalent concept of hot-nodes, which are formally defined later. Thus, SMART-3D can be applied to higher dimensional spaces. While this paper focuses specifically on $\mathbb{R}^3$ space, it could be extended to $\mathbb{R}^n$ spaces in general.

\begin{figure*}[t]
    \centering
    \subfloat[The current path in LRZ becomes invalid. Thus, the CPR is identified and risky nodes in CPR are pruned, forming multiple disjoint subtrees.]{
        \includegraphics[width=0.32\textwidth]{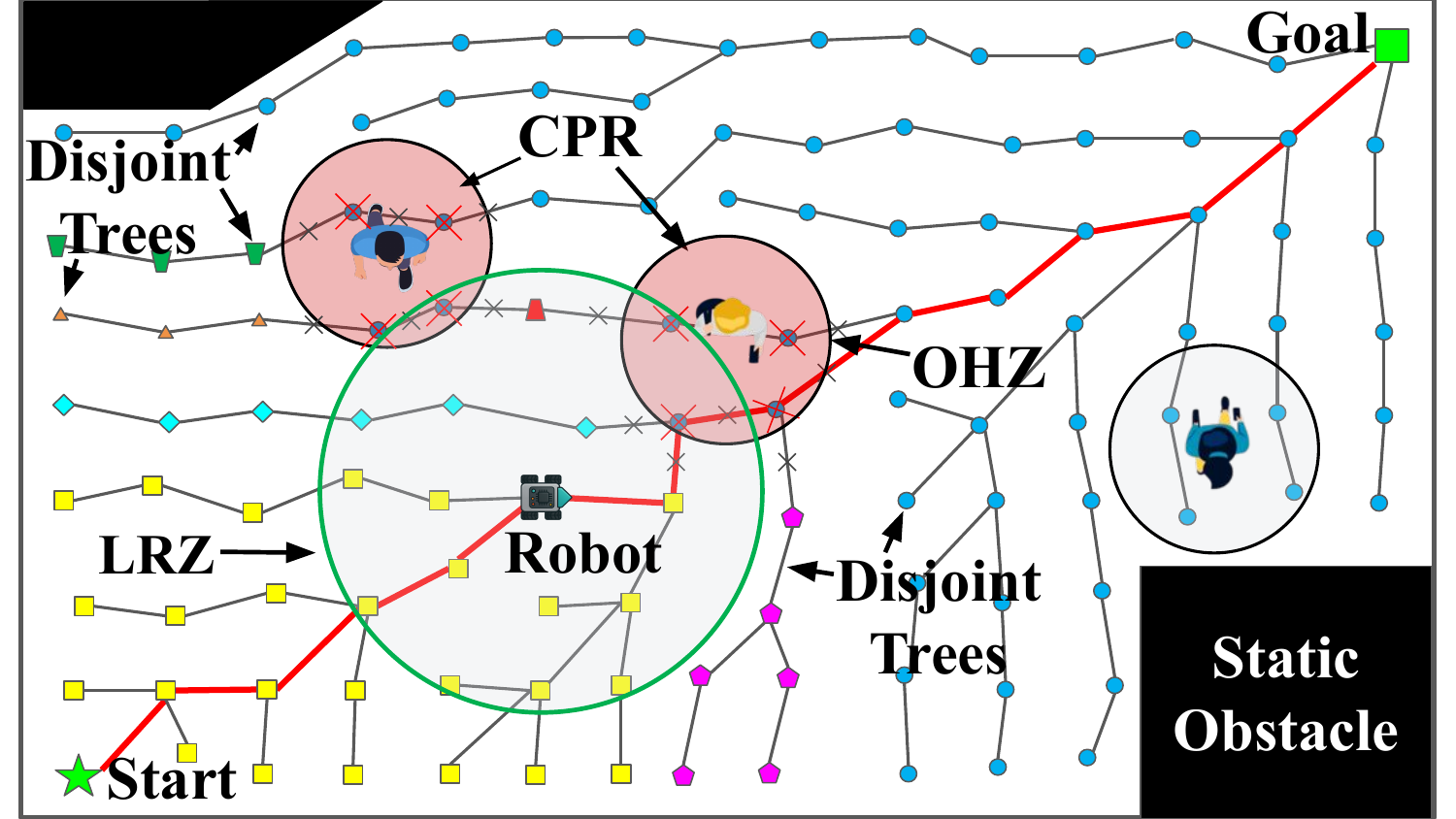}\label{fig:example_part1}}\vspace{0pt}\hspace{-8pt}\quad
    \centering
    \subfloat[The search for hot-nodes starts in LSR centered at the closest pruned path node to the robot. Then, the hot-node with the highest utility is selected.]{
         \includegraphics[width=0.32\textwidth]{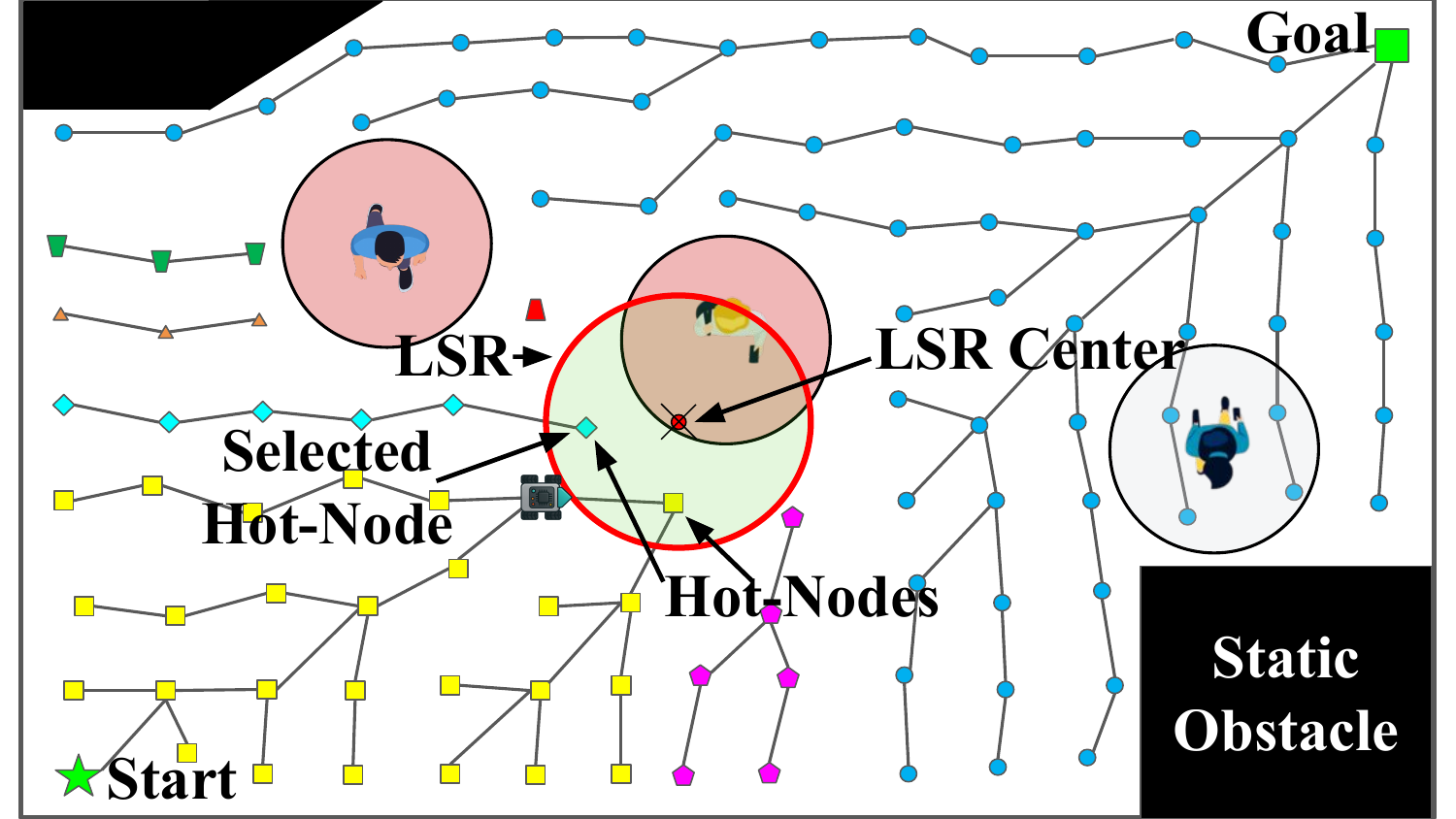}\label{fig:example_part2}}\hspace{-8pt}\quad
    \centering
    \subfloat[The tree-repair process is started by connecting the selected hot-node to its neighboring nodes that belong to different disjoint subtrees.
    ]{
         \includegraphics[width=0.32\textwidth]{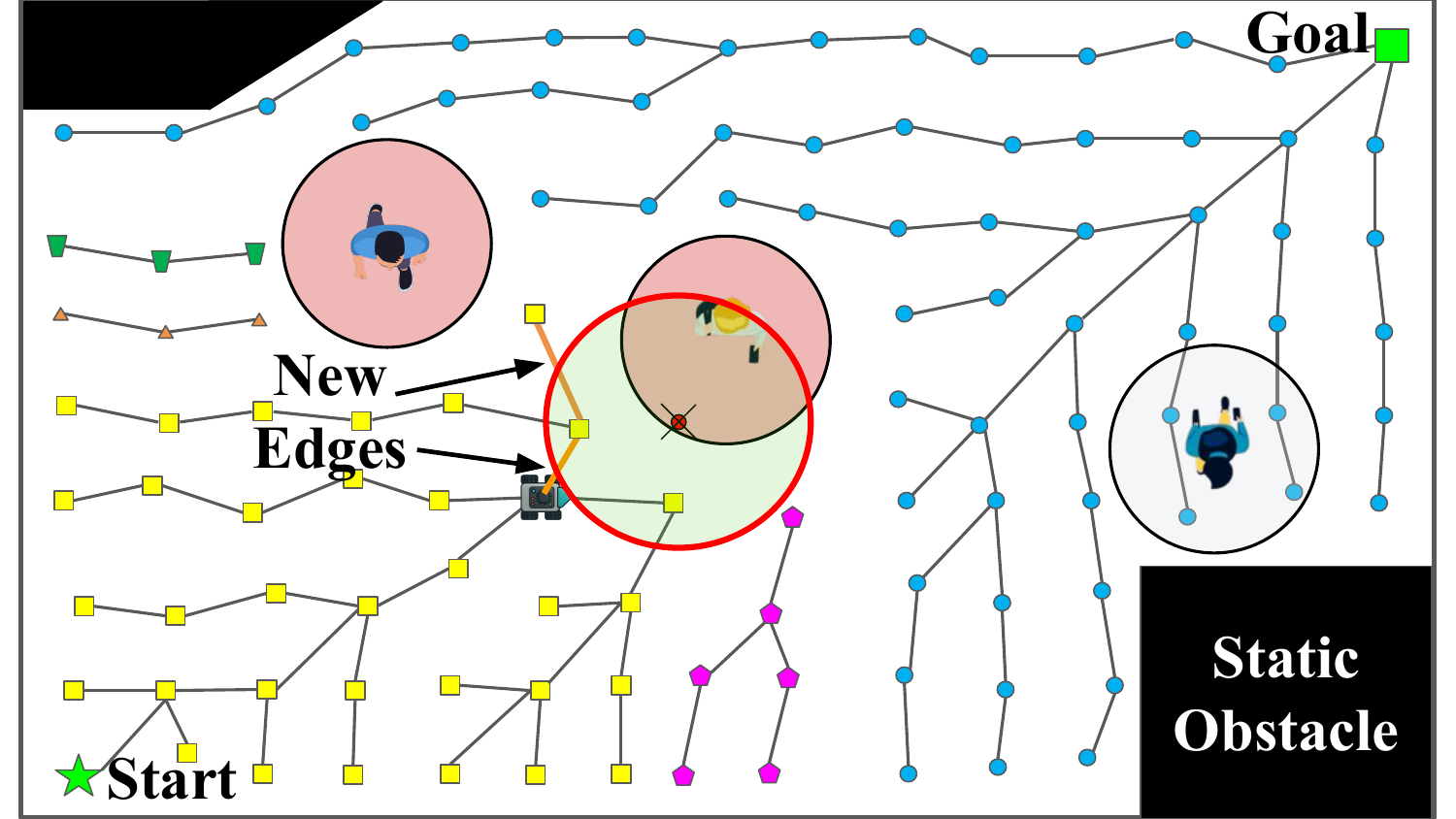}\label{fig:example_part3}}\hspace{-8pt}\\

    \centering
    \subfloat[The hot-node with the next highest utility is selected.]{
        \includegraphics[width=0.32\textwidth]{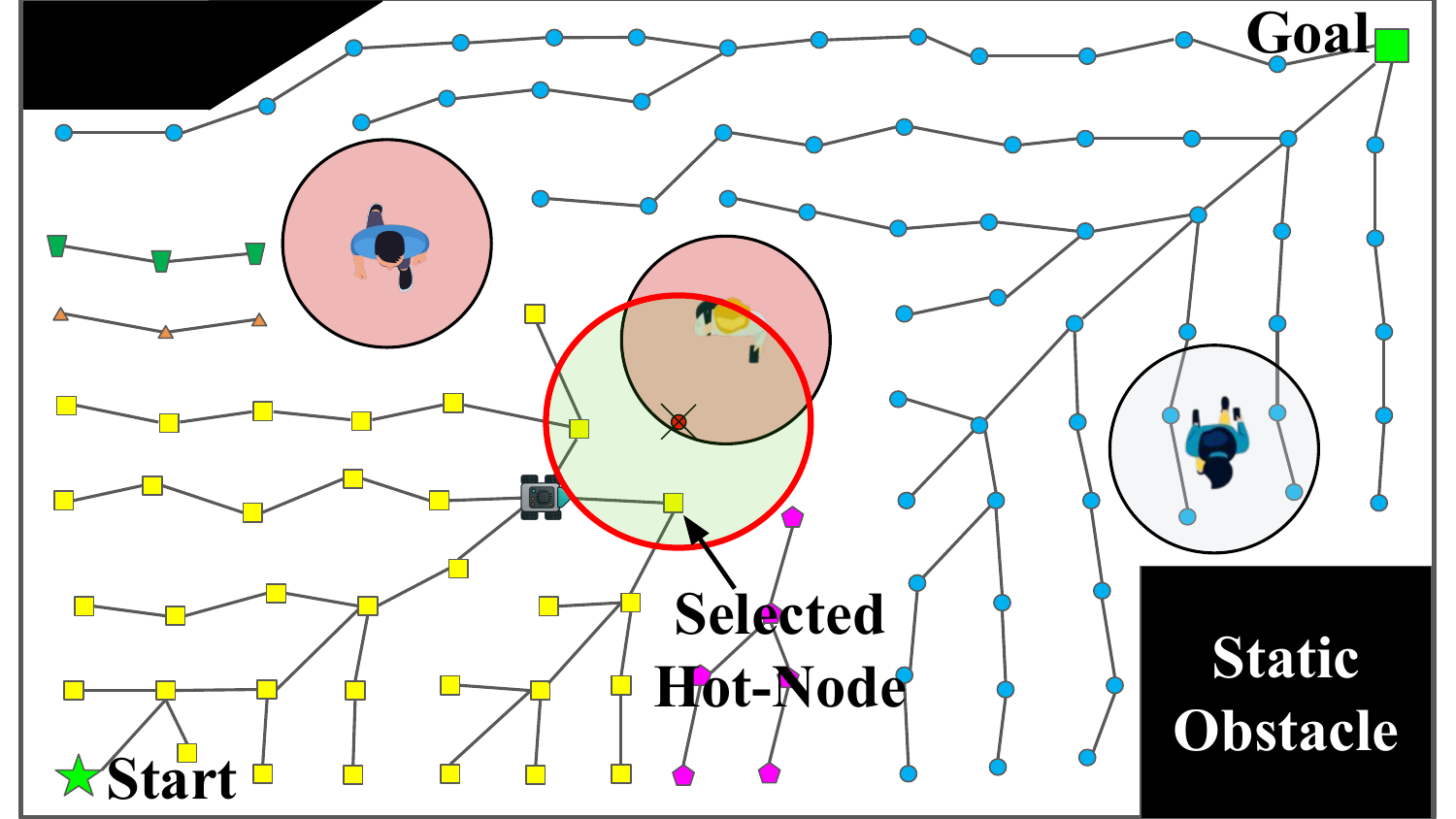}\label{fig:example_part4}}\vspace{0pt}\hspace{-8pt}\quad
    \centering
    \subfloat[The tree is further repaired at the selected hot-node. No more hot-nodes exist in LSR while the goal-rooted subtree $\mathcal{T}_0$ is still disconnected.]{
         \includegraphics[width=0.32\textwidth]{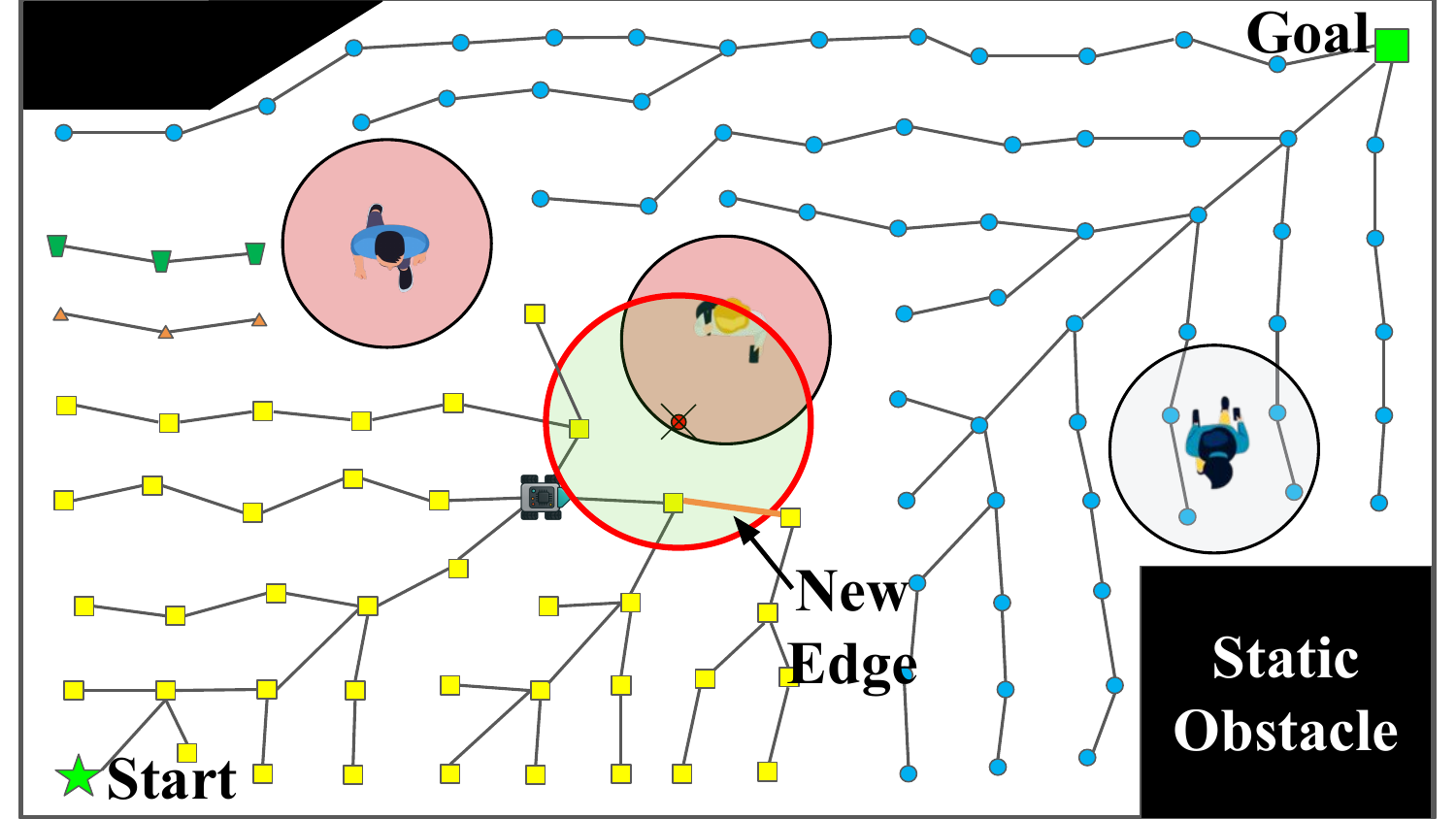}\label{fig:example_part5}}\hspace{-8pt}\quad
     \centering
    \subfloat[The LSR is expanded by increasing its radius. The hot-nodes are searched in the expanded LSR and the one with the highest utility is selected.]{
         \includegraphics[width=0.32\textwidth]{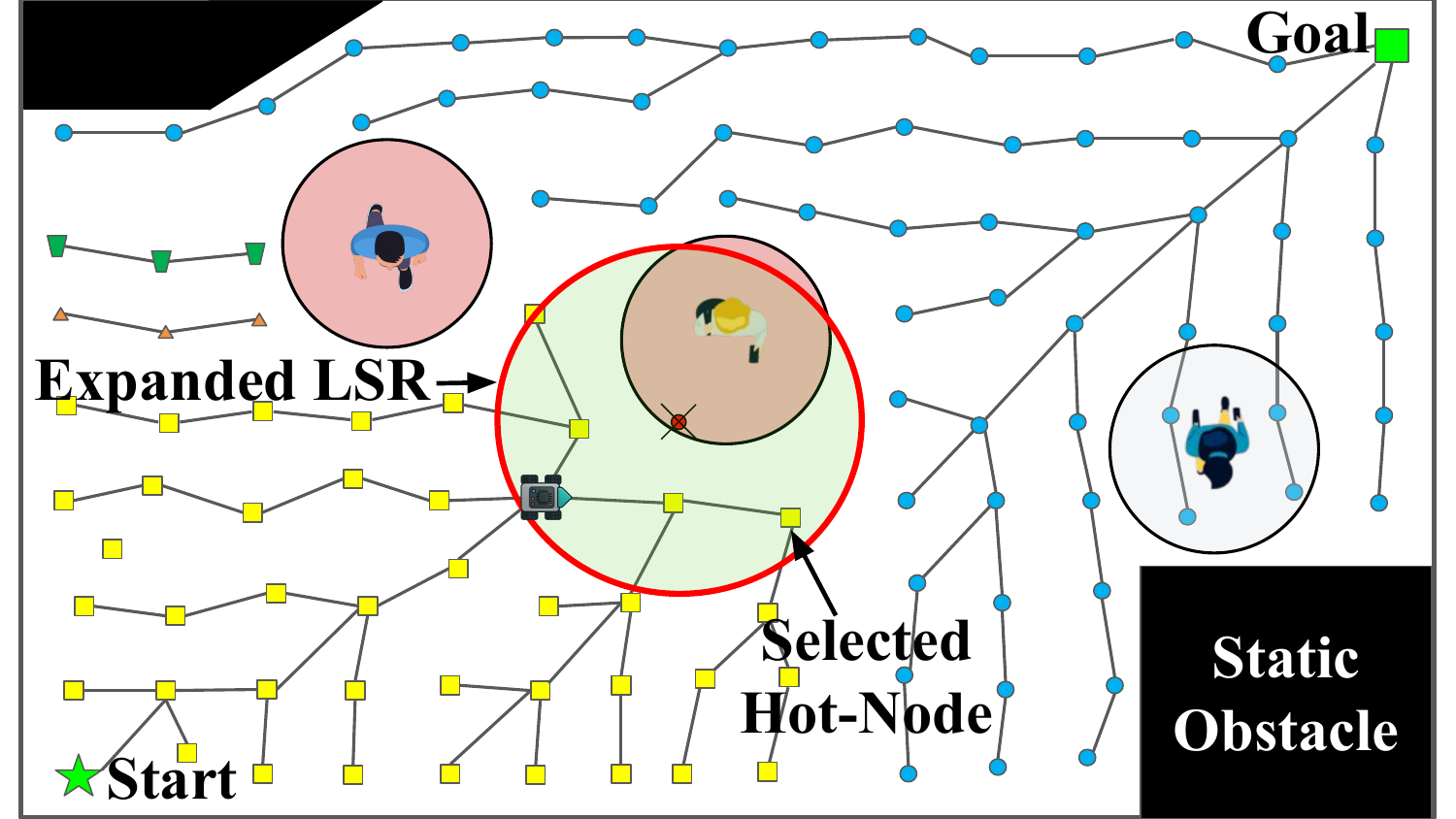}\label{fig:example_part6}}\hspace{-8pt}\\

    \centering
    \subfloat[The tree is further repaired at the selected hot-node. Now, the goal-rooted subtree $\mathcal{T}_0$ is connected.]{
        \includegraphics[width=0.32\textwidth]{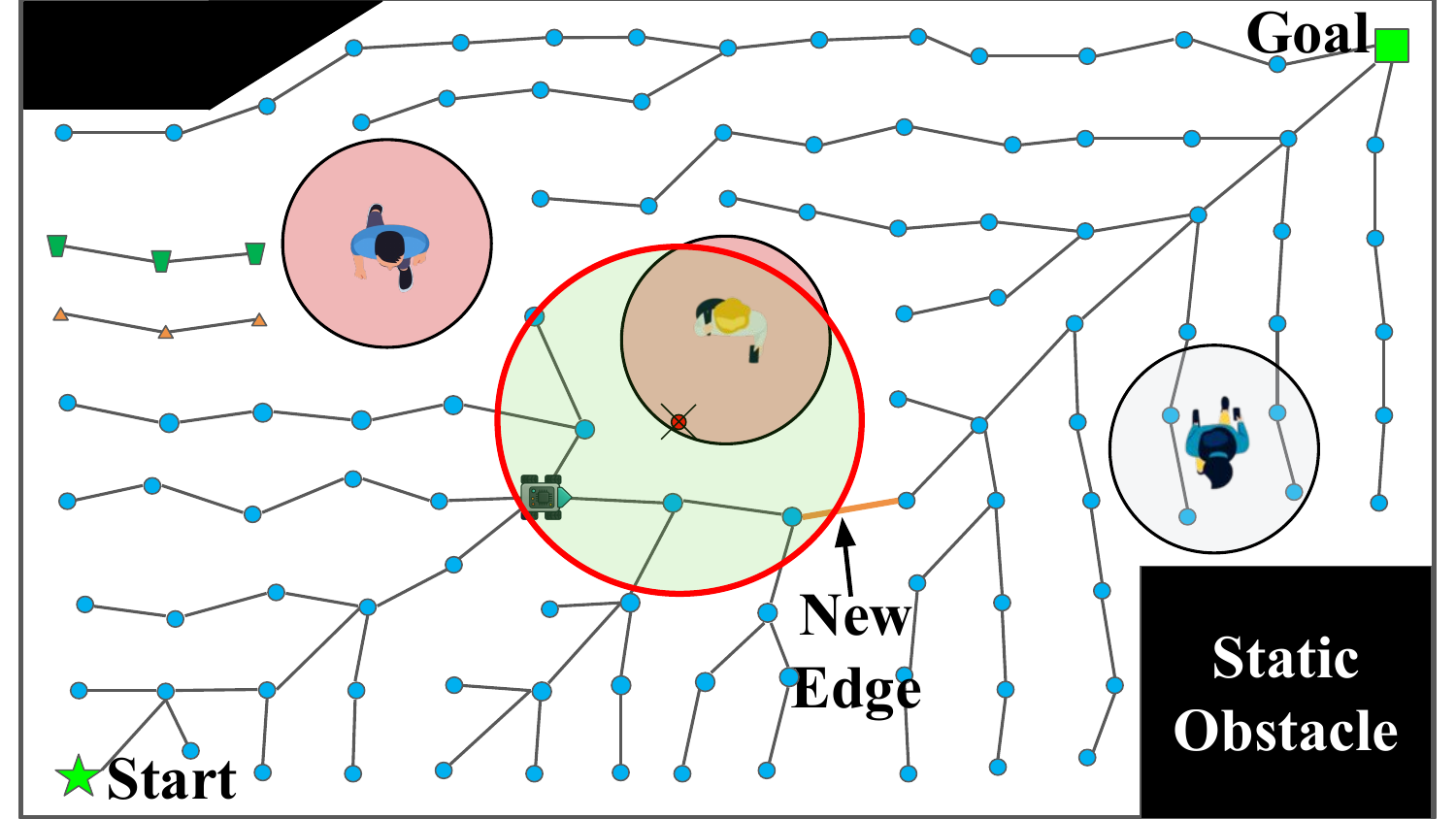}\label{fig:example_part7}}\vspace{0pt}\hspace{-8pt}\quad
    \centering
    \subfloat[The repaired tree is optimized via a rewiring cascade, updating the cost-to-go of its nodes.]{
         \includegraphics[width=0.32\textwidth]{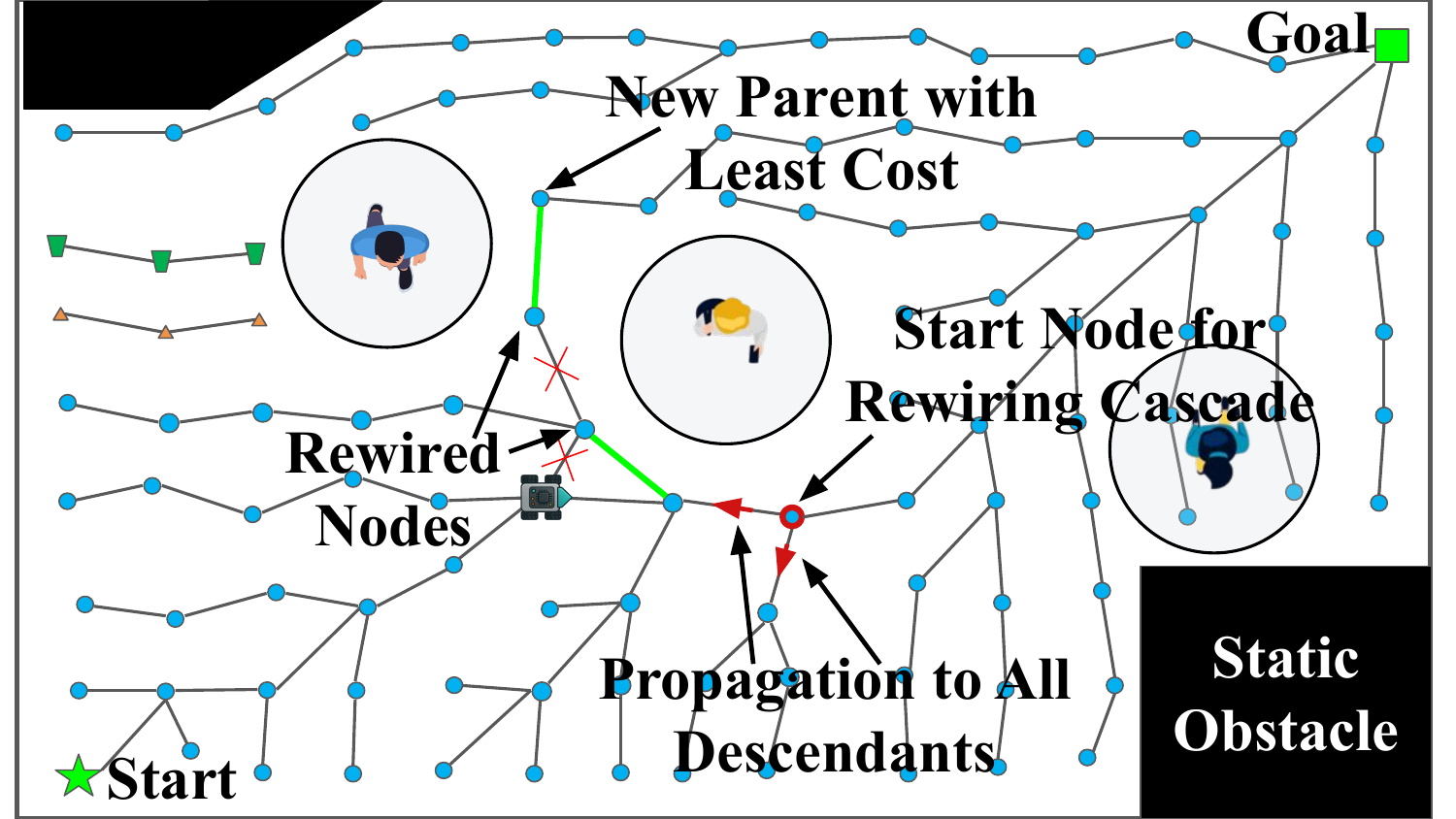}\label{fig:example_part8}}\hspace{-8pt}\quad
    \centering
    \subfloat[Finally, the path is replanned. Then, all pruned nodes are added back and any disjoint subtrees left are merged to form a single morphed tree $\mathcal{T}$.]{
         \includegraphics[width=0.32\textwidth]{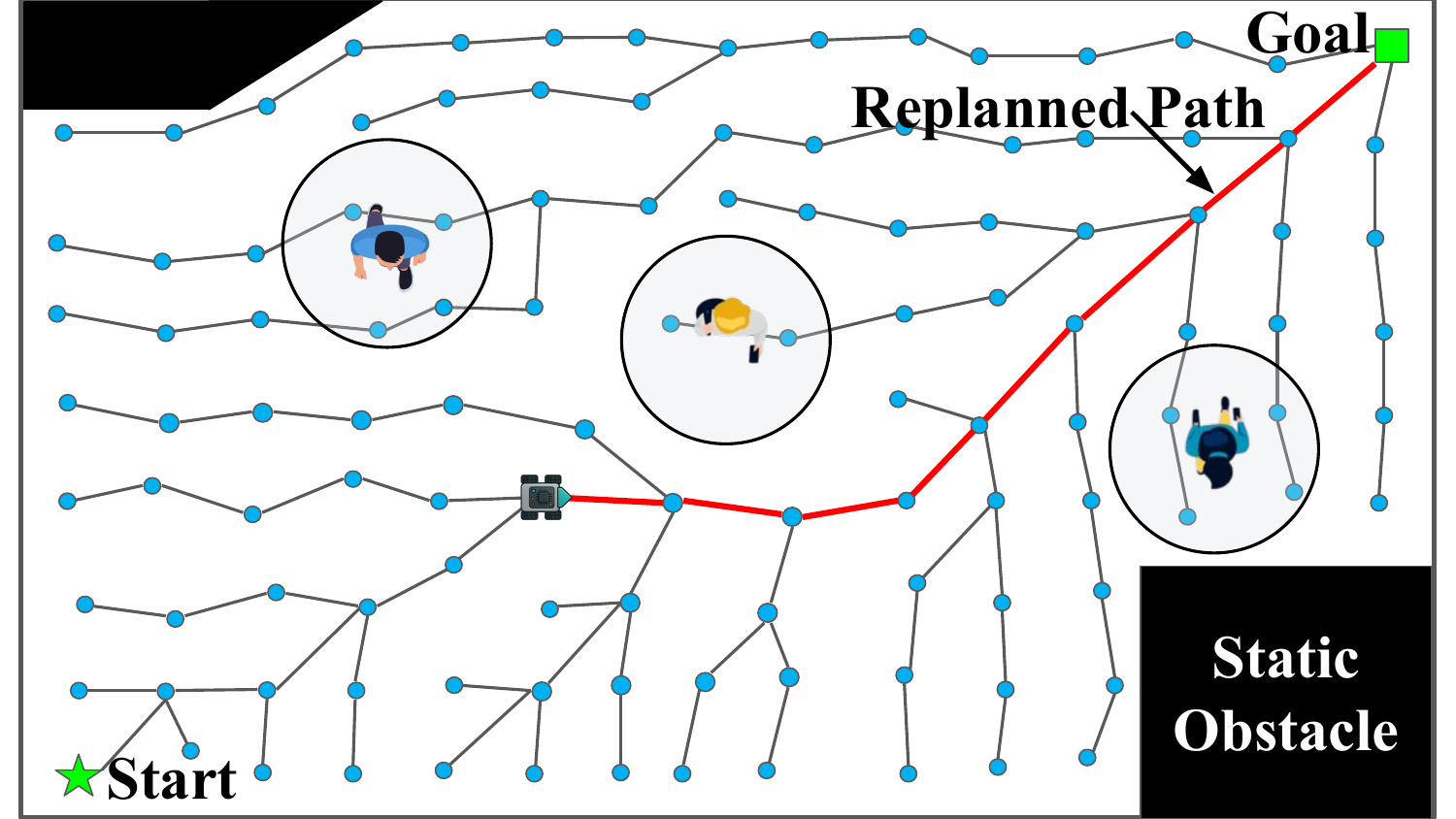}\label{fig:example_part9}}\hspace{0pt}\\
          \caption{Illustration of replanning done by SMART-3D: a) tree-pruning and disjoint subtree creation, and b)-i) tree-repair and replanning.}\label{fig:example}
          \vspace{0pt}
 \end{figure*}

Now we present the details of the algorithm. Let \(\mathcal{X} \subset \mathbb{R}^3\) be a space containing static and dynamic obstacles. Let \(\mathcal{X}_{N} \subset \mathcal{X}\) be the configuration space free of static obstacles. Let \(\mathcal{O} = \{O_i : i = 1, 2, \ldots, m\}\) be the set of \(m\) dynamic obstacles. For ease and efficiency of collision checking, we assume that all obstacles are spherical, where \(r_i \in \mathbb{R}^+\) is the radius of obstacle \(O_i \in \mathcal{O}\). The position and speed of $O_i$ at time \(t \in \mathbb{R}^+\) are denoted by \(x_i(t) \in \mathcal{X}_N\) and \(v_i(t) \in \mathbb{R}^+\), respectively. Let \(R\) be a spherical robot of radius \(r_R \in \mathbb{R}^+\), whose position and speed at time \(t\) are denoted by \(x_R(t) \in \mathcal{X}_N\) and \(v_R(t) \in \mathbb{R}^+\), respectively. Let \((x_s, x_g)\) denote the start and goal positions.

\subsection{Initialization}
To initialize, a RRT$^*$~\cite{rrtstar} tree $\mathcal{T}^0 \equiv (\mathcal{N}^0, \mathcal{E}^0)$ is constructed by considering only the static obstacles in the environment, where $(\mathcal{N}^0, \mathcal{E}^0)$ denote the pair of its node and edge sets. $\mathcal{T}^0$ is rooted at the goal $x_g$. In order to achieve the best replanning performance, $\mathcal{T}^0$ should cover \(\mathcal{X}_{N}\) with sufficient density such that new optimal paths could be found with ease without the need for further sampling and expanding the tree. Each node in the tree maintains a data structure containing information about its a) position, b) parent node, c) children nodes, d) subtree index, e) cost-to-go to the root, and f) status as active or pruned. Initially, all nodes are active and have the same index. An initial path $\sigma^0$ is found in \(\mathcal{X}_{N}\) using $\mathcal{T}^0$. 

\vspace{0pt}
\subsection{Tree-Pruning}
\label{prune}
As the robot moves, $\sigma^0$ could be blocked by dynamic obstacles and become infeasible. Thus, in order to replan and find a safe path, it is required to prune the nodes and edges which are risky or colliding with these obstacles. To save computation effort and since the obstacles far away from the robot do not pose an immediate collision risk, the path feasibility checking and tree-pruning are done in a local neighborhood of the robot. This process is described below.

First, we define a Local Reaction Zone (LRZ) around the robot and an Obstacle Hazard Zone (OHZ) around each dynamic obstacle. While for 2D case the LRZ and OHZs are circular disks, for 3D case they are spheres centered at the robot.  The LRZ represents the region reachable by the robot in a certain reaction time-horizon. The OHZ is a high-risk region around a given obstacle. Since obstacles could have unpredictable movements, the OHZ is defined as a sphere with the obstacle's radius plus its speed times a risk time-horizon. The LRZ and OHZ are defined~\cite{smart} as follows:

\begin{defn}[LRZ]\label{define:LRZ}
The \textup{LRZ} for a robot $\mathcal{R}$ is defined as $\textup{LRZ}_{\mathcal{R}}(t)=\{x\in \mathcal{X}_N:||x-x_{\mathcal{R}}(t)||\leq $ $v_{\mathcal{R}}(t)\times T_{RH}\}$, where $T_{RH} \in \mathbb{R}^+$ is the robot reaction time-horizon. 
\end{defn}

\begin{defn}[OHZ]\label{define:OHZ}
The \textup{OHZ} for an obstacle $O_i \in \mathcal {O}$ is defined as $\textup{OHZ}_{i}(t) =\{x\in \mathcal{X}_N: ||x-x_{i}(t)||\leq v_i(t)\times T_{OH} + r_{i} + r_{\mathcal{R}} \}$, where $T_{OH} \in \mathbb{R}^+$ is the obstacle risk time-horizon.
\end{defn}

Note: For convenience, the robot radius $r_{\mathcal{R}}$ is added to the OHZ to then treat the robot as of point size. 

The nearby obstacles whose OHZs intersect with the robot's LRZ at a certain time can pose a collision risk to the robot. Thus, the nodes and edges in these OHZs must be pruned as they cannot be used to plan a new path. The union of all these OHZs form a Critical Pruning Region (CPR) as defined below.

\begin{defn}[CPR]\label{define:CPR}
Let $\mathcal{D}$ denote all obstacles that are intersecting with the robot's \textup{LRZ} and pose danger to the robot such that $\mathcal{D}=\{O_i \in \mathcal{O}: \textup{LRZ}_{\mathcal{R}}(t)\cap \textup{OHZ}_i(t) \neq \emptyset\}$. Then, the \textup{CPR} is defined as
\begin{equation}
\textup{CPR}(t)= \bigcup_{\mathcal{D}} \textup{OHZ}_i(t).
\end{equation}
\end{defn}

The robot constantly checks the feasibility of the portion of its current path in LRZ. If the current path in LRZ is invalid, then all nodes and edges that fall in the entire CPR are pruned. If a node falls in CPR, then it is pruned along with its connecting edges. If an edge falls in CPR although it is connected to two safe nodes outside CPR, then it is pruned but its end nodes are kept. Figure \ref{fig:example_part1} shows an illustrative example of tree-pruning in CPR when the robot's current path in its LRZ is blocked by an obstacle. The tree-pruning process could result in creation of multiple disjoint subtrees.

\begin{defn}[Disjoint Subtree]\label{define:dissub}
A disjoint subtree is a broken portion of the current tree $\mathcal{T}$ whose root is either the goal or a child of a pruned node or edge. 
\end{defn}

Let $\mathcal{T}\equiv (\mathcal{N},\mathcal{E})$ be the current tree. Suppose the tree-pruning process breaks $\mathcal{T}$ into $K\in \mathbb{N}^+$ different disjoint subtrees denoted by $\mathcal{T}_0,\mathcal{T}_1,...\mathcal{T}_{K-1}$. Here  $\mathcal{T}_0\equiv (\mathcal{N}_0,\mathcal{E}_0)$ is the subtree rooted at the goal. Let $\mathcal{N}^a \in \mathcal{N}$ and $\mathcal{N}^p \in \mathcal{N}$ be the sets of all alive and pruned nodes, respectively, such that $\mathcal{N}=\mathcal{N}^a \bigcup \mathcal{N}^p$. 

Each node $n \in \mathcal{N}^a$ stores its own subtree index. When a node is pruned, its children (none, one or many) become the roots of their own disjoint subtrees and the subtree indices are updated for all their descendants. Similarly, when an edge is pruned, the orphaned node becomes the root of its own disjoint subtree and the subtree index is updated for all its descendants. 

\vspace{0pt}
\subsection{Tree Repair}
The objective of tree-repair is to stitch together all disjoint subtrees formed after the tree-pruning process to form a single morphed tree which can then be used to obtain a new path to the goal. Since the node density of $\mathcal{T}$ is sufficiently high in the free space outside CPR, its easy to repair the disjoint subtrees into a single tree by making simple reconnections. Thus, it is required to identify nodes which can be easily connected with nodes of other disjoint subtrees. 

As compared to hot-spots~\cite{smart} in SMART, this paper uses the concept of hot-nodes, which does not require a grid decomposition, thus making the algorithm computationally efficient and extendable to high-dimensional spaces. The tree-repair process works by searching for hot-nodes and incrementally making reconnections between disjoint subtrees at the hot-nodes with the highest utilities.

\begin{defn}[Neighbor]\label{define:neighbor}
A node $n_j \in \mathcal{N}^a$ is said to be a neighbor of node $n_i \in \mathcal{N}^a$ if $x_{n_j}\in \mathcal{B}(x_{n_i},r)$, where $x_{n_i}\in \mathcal{X}_N$ and $x_{n_j} \in \mathcal{X}_N$ are the positions of $n_i$ and $n_j$, respectively, and $\mathcal{B}(x_{n_i},r)$ is  a ball of radius $r$ centered at $n_i$.
\end{defn}

\begin{defn}[Eligible Neighbor]\label{define:eligibleneighbor}
A neighbor $n_j \in \mathcal{N}^a$ of node $n_i \in \mathcal{N}^a$ is said to be an eligible neighbor if it belongs to a different disjoint subtree than $n_i$ and the nodes $n_i$ and $n_j$ could be connected by a feasible edge.
\end{defn}

\begin{defn}[Hot-Node]\label{define:hotnode}
A node $n \in \mathcal{N}^a$ is said to be a hot-node if it has at least one eligible neighbor.
\end{defn}

\vspace{6pt}
\subsubsection{Search for Hot-Nodes}\label{search}
The search for hot-nodes begins around the damaged path. Therefore, we define a Local Search Region (LSR) centered at the pruned node on the damaged path which is nearest to the robot.

\begin{defn}[LSR]\label{define:LSR}
Let $\hat{n} \in \mathcal{N}^p$ be the pruned path node closest to the robot. Let $x_{\hat{n}}$ be the location of $\hat{n}$. Let $r_{s}$ be the search radius. Then, the \textup{LSR} is defined as $\textup{LSR}(r_s) = \{x\in \mathcal{X}_N: ||x-x_{\hat{n}}||\leq r_s\}$.
\end{defn}

At each replanning incident, the search radius $r_s$ is initialized to be a user-specified radius, $r_{s_0}$. Then, the LSR is searched and all hot-nodes are identified. Fig. \ref{fig:example_part2} shows an example of hot-node search in LSR. If there are no hot-nodes in LSR, then the LSR is expanded by multiplying $r_s$ by an expansion factor $\lambda>1$ (Fig. \ref{fig:example_part6}). This process is repeated until at least one hot-node is found. 

\vspace{6pt}
\subsubsection{Ranking of Hot-Nodes} \label{ranking}

After all hot-nodes are detected in the LSR, they are ranked by utility as defined as follows. 

\begin{defn}[Utility]\label{define:utility}
The utility of a hot-node $n$ is computed as follows:
\begin{equation}
\mathcal{U}(n) = 
\begin{cases}
\frac{1}{
\|x_{\mathcal{R}} - x_n\|_2 + \|x_{n} - x_{n'}\|_2 + g(n')}, & \text{if } n' \in \mathcal{N}_0 \\
\frac{1}{
\|x_{\mathcal{R}} - x_n\|_2 + \|x_{n} - x_{n'}\|_2 + \|x_{n'} - x_g\|_2}, & \text{if } n' \notin \mathcal{N}_0
\end{cases}
\label{eq:utility}
\end{equation}
 where $n'$ is the nearest eligible neighbor of $n$; $x_n$ and $x_{n'}$ are the positions of $n$ and $n'$, respectively; and $g(n')$ returns the travel cost from $n'$ to the goal on $\mathcal{T}_0$. 
\end{defn}

Thus, if the nearest neighbor $n'$ of hot-node $n$ belongs to the goal-rooted tree $\mathcal{T}_0$, then the actual cost-to-go is used as the distance to the goal, otherwise a heuristic cost is used.

\RestyleAlgo{ruled}
\LinesNumbered
\begin{algorithm}[t]
\footnotesize
$\{\mathcal{T}^0,\sigma^0\} \leftarrow$ \textbf{RRT*}($x_{s},x_{g},\mathcal{X}_{N}$)\tcp*[l]{initialization}
$t=t_0$,  $\mathcal{T}= \mathcal{T}^0$, $\sigma=\sigma_{0}$\; 
\While (\tcp*[h]{goal unreached}) {$x_{\mathcal{R}} \neq x_{g}$}{
    $t \leftarrow$ \textbf{UpdateClock}()\;
    $\{x_i(t),v_i(t)\}_{i=1,..m} \leftarrow$ \textbf{UpdateObstacleState}()\;
    $\{x_{\mathcal{R}}(t),v_{\mathcal{R}}(t) \} \leftarrow$ \textbf{UpdateRobotState}()\;
    \eIf{\textbf{ValidatePath}($\sigma, \textup{LRZ}_{\mathcal{R}}(t), \{\textup{OHZ}_i(t)\}_{i=1,..m})$}{
    \textbf{Navigate}($\sigma$)\;
    }{
        $\sigma \leftarrow$ void\tcp*[l]{invalid path}
        
        $\{\mathcal{T}_0,...\mathcal{T}_{K-1}\} \leftarrow$ \textbf{TreePruning} ($\mathcal{T}$,\textup{CPR}(t))\tcp*[l]{\ref{prune}}
        
        $r_s \leftarrow r_{s_{0}}, \mathcal{H} \leftarrow \emptyset$, $\mathcal{N}_{s} \leftarrow \emptyset$\tcp*[l]{initialize repair}
        
            \While {$\mathcal{T}_0$ is not reachable from $x_{\mathcal{R}}$}{
                \eIf(\tcp*[h]{informed tree-repair}){$r_s < {r_{s_{max}}}$}{
                    $r_s \leftarrow r_s \times \lambda$\;
                    $\mathcal{H} \leftarrow$ \textbf{HotNodeSearch}(\textup{LSR}($r_s$))\tcp*[l]{\ref{search}}
                    $\mathcal{U} \leftarrow$ \textbf{ComputeUtility}($\mathcal{H}$)\tcp*[l]{\ref{ranking}} 
                   $\{{\mathcal{N}}_s,{\mathcal{T}}_0 \} \leftarrow$ \textbf{TreeReconnection}($\mathcal{H}, \mathcal{U}$)\tcp*[l]{\ref{repair}}
                }(\tcp*[h]{standard tree-repair}){
                    $x_{rand} \leftarrow$ \textbf{SampleFree()}\;
                    Join $x_{rand}$ to all nearby reachable subtrees\;
                    \lIf{$x_{rand} \in \mathcal{T}_0$}{$\textbf{Rewire}(x_{rand})$}
                                        
                }
            }
            $\mathcal{T}_0 \leftarrow\textbf{TreeOptimization}(\mathcal{T}_0,\mathcal{N}_s)$\tcp*[l]{\ref{tree-opt}}
            $\sigma \leftarrow \textbf{PathSearch}(x_{\mathcal{R}}, x_g, \mathcal{T}_0)$\tcp*[l]{\ref{path}}
            Add $\mathcal{N}^p$ and disjoint subtrees to $\mathcal{T}_0$ to get a single tree $\mathcal{T}$\;
            
    }
}
\caption{SMART-3D}
\label{alg:SMART} 
\end{algorithm}
\setlength{\textfloatsep}{12pt}

\begin{figure*}[t]
    \centering
     \subfloat[2D Scenario with $15$ dynamic obstacles. The initial path (green) connects the start (blue) and goal (red) points.  
     ]{
         \includegraphics[width=0.37\linewidth]{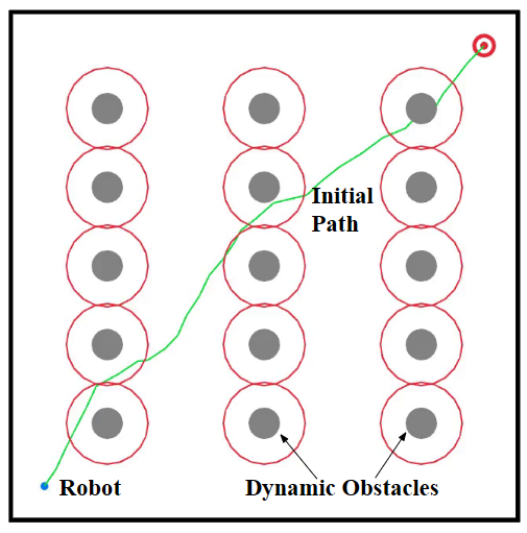}\label{fig:2d-scenario-15-obs}}\hspace{50pt} 
    \subfloat[3D Scenario with 100 randomly distributed dynamic obstacles. The initial path (green) connects the start (blue) and goal (red) points.]{
         \includegraphics[width=0.37\linewidth]{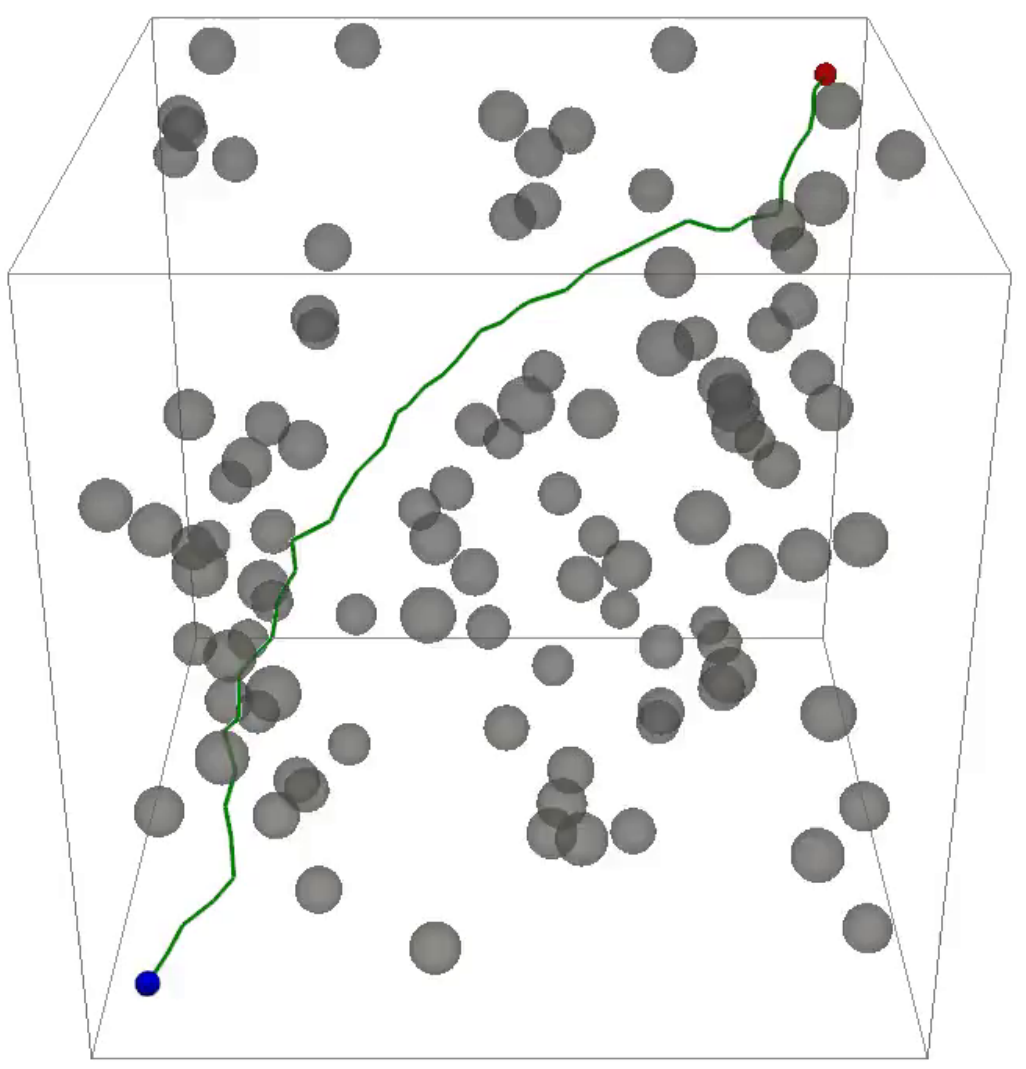}\label{fig:3d-scenario-100-obs}}\hspace{-8pt}
    \caption{Examples of 2D and 3D scenarios. 
    Initial path is found for both these scenarios using RRT* without considering dynamic obstacles.}
    \label{fig:3d-scenarios} \vspace{0pt}
\end{figure*}

\vspace{6pt}
\subsubsection{Tree-reconnections}\label{repair} Once the hot-nodes in LSR are identified and ranked, tree reconnections are done. This consists of the following steps:
\begin{itemize}
\item [a.] Pick the hot-node in LSR with the highest utility (Figs. \ref{fig:example_part2}, \ref{fig:example_part4}). If no hot-nodes are left in LSR, then go back to search in expanded LSR (Fig. \ref{fig:example_part6}).
\item [b.] Pick the nearest eligible neighbor of the hot-node and connect it (Figs. \ref{fig:example_part3}, \ref{fig:example_part5}, \ref{fig:example_part7}).
\item [c.] Update the  parent-child relationships and subtree indices (Figs. \ref{fig:example_part3}, \ref{fig:example_part5}, \ref{fig:example_part7}) as follows: 1) if any node from the above node-pair belongs to $\mathcal{T}_0$, then this node is set as the parent and all nodes of the other connected subtree are assigned the index of $\mathcal{T}_0$ and their cost-to-go are updated, or 2) if none of the nodes in the above node-pair belongs to $\mathcal{T}_0$, then one of them is set as the parent and all nodes of the other connected subtree are assigned its index. 
\item [d.] Update the hot-nodes in LSR and their utilities.
\item [e.]  Check if $\mathcal{T}_0$ is reachable from $x_{\mathcal{R}}$ (Fig. \ref{fig:example_part7}). If true, then a path can be found to the goal; otherwise return to step a above.
\end{itemize}

Note: If the entire space is exhausted by tree-repair
and no path is found using alive samples, then random
sampling and repairing is done for probabilistic completeness.

\subsection{Tree-Optimization}
\label{tree-opt} 
During tree-repair, several subtrees were merged with $\mathcal{T}_0$. These subtrees are optimized using a rewiring cascade starting from all subtree nodes $\mathcal{N}_s$ connected to $\mathcal{T}_0$ (Fig. \ref{fig:example_part8}).

\subsection{Path Search}
\label{path}
After the merged tree is optimized, an updated path $\sigma$ is found using this tree (Fig. \ref{fig:example_part9}). 
After we find a new path, all nodes in $\mathcal{N}^p$ and the roots of the remaining disjoint subtrees are reconnected with $\mathcal{T}_0$ and a single tree $\mathcal{T}$ is formed.

Algorithm \ref{alg:SMART} shows the workings of SMART-3D. The pseudocode remains the same as that of \cite{smart}, with the exception that the hot-spot search is replaced with a hot-node search.

\begin{table}
\centering
\caption{Parameters for 2D and 3D Scenarios}
\begin{tabular}{l|cc}
\hline
\textbf{Parameter} & \textbf{2D Scenario} & \textbf{3D Scenario} \\
\hline
Risk Time-Horizon ($T_{OH}$) & $0.4\,\text{s}$ & $0.4\,\text{s}$ \\
Reaction Time-Horizon ($T_{RH}$) & $1\,\text{s}$ & $1\,\text{s}$ \\
LSR Initial Radius ($r_{s_0}$) & $1\,\text{m}$ & $1\,\text{m}$ \\
LSR Expansion Factor ($\lambda$) & $1.5$ & $1.5$ \\
LSR Max Radius ($r_{s_{max}}$) & $10\,\text{m}$ & $10\,\text{m}$ \\
Neighborhood Radius ($r$) & $1.7\,\text{m}$ & $1.7\,\text{m}$ \\
Initial RRT* Steering Range & $1\,\text{m}$ & $1\,\text{m}$ \\
Initial RRT* Iterations & $2500$ & $20000$ \\
\hline
\end{tabular}
\vspace{5pt}
\label{tab:smart3d_params}
\end{table}

\vspace{0pt}
\section{Results}
This section presents the performance results of SMART-3D via Monte carlo simulations of a robot traversing in open environments with several dynamic obstacles.  The robot is considered to be holonomic with radius $r_{\mathcal{R}}=0.5$m and speed $v_{\mathcal{R}}=4$m/s. For simplicity, we treat the robot as a point and add its radius to the OHZs for collision checking. The results are shown for 2D and 3D environments. Table \ref{tab:smart3d_params} lists all parameters for the 2D and 3D simulations. Each simulation trial has a timestep of $0.1s$. If any replanning event takes longer than $0.1$s, the whole trial is marked as failed. However, this is rare and typically occurs in the case where multiple obstacles surround the robot such that there is no solution, or the only solution is through an extremely narrow opening between obstacles. Also, if a collision occurs, then the trial is marked as failed. For successful trials, the following performance metrics are used for validation: 
\begin{itemize}
    \item \textit{Replanning time}: Time taken by the robot to replan its path when its current path is broken. 
    \item \textit{Success rate}: Fraction of successful trials where the robot reached the goal without collision.
    \item \textit{Travel time}: Time taken by the robot to travel from the start to the goal on a successful trial.
\end{itemize}


\begin{figure*}[!t]
  \centering
  \subfloat[\hspace{0pt} Success rate over $100$ trials]{%
    \includegraphics[width=0.33\textwidth]{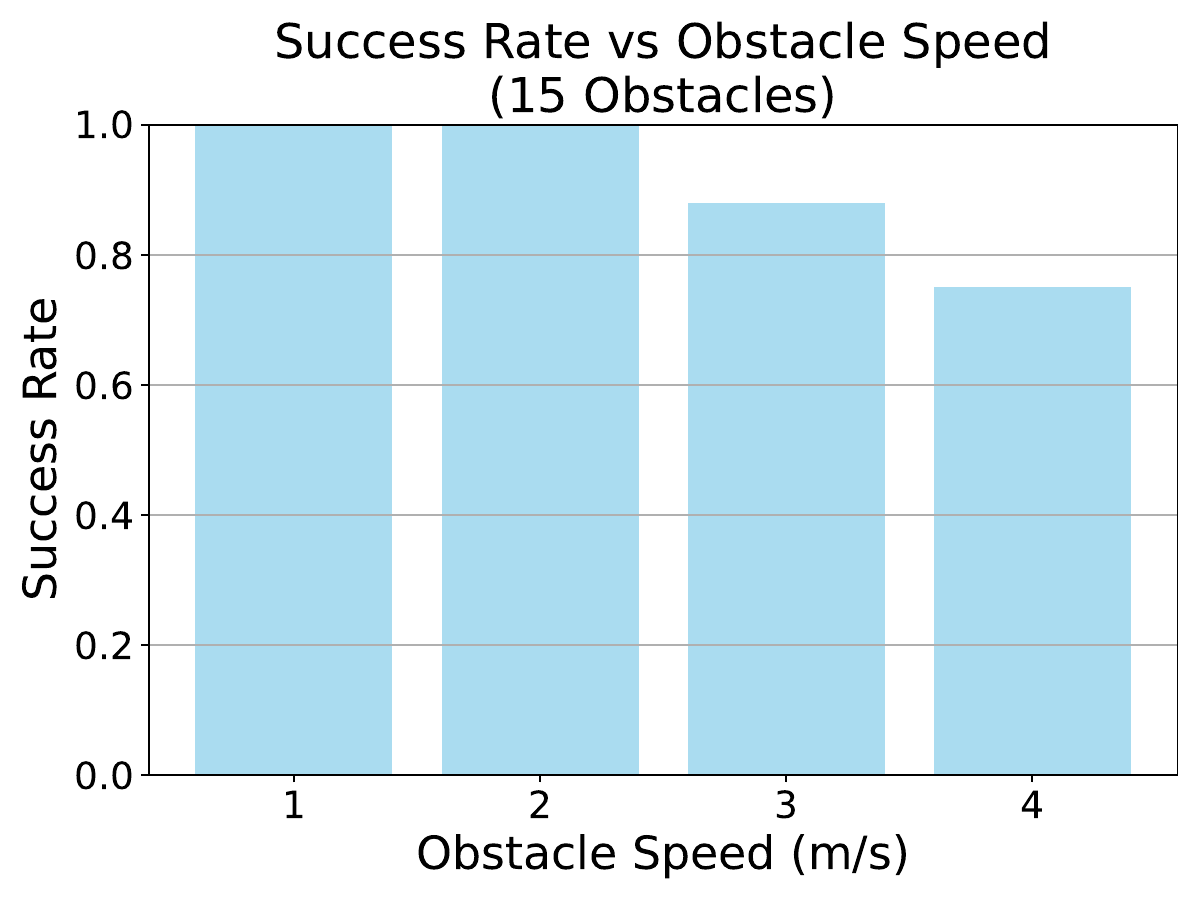} \label{fig:2D-a}%
  }\hfill
  \subfloat[\hspace{0pt} Average replanning time per trial]{%
    \includegraphics[width=0.33\textwidth]{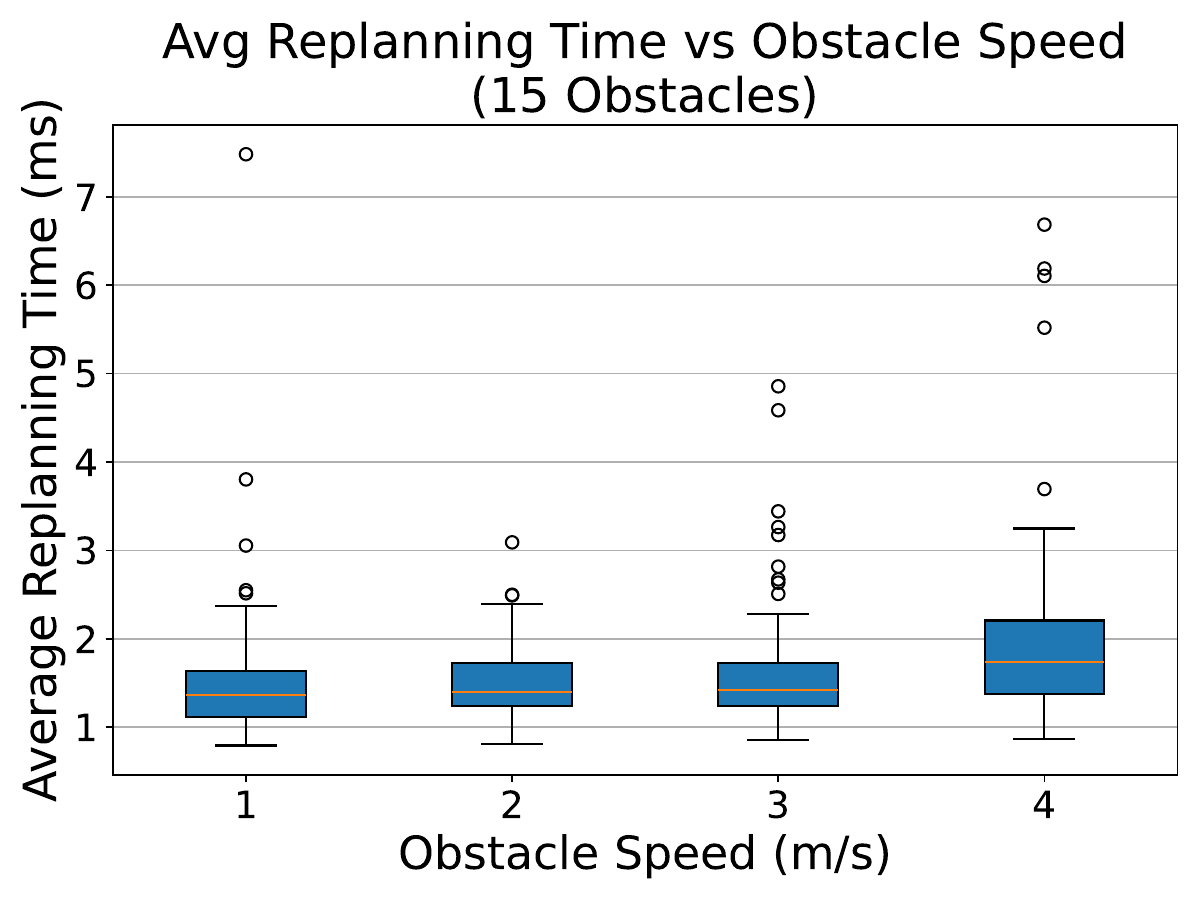}\label{fig:2D-b}%
  }\hfill
  \subfloat[\hspace{0pt} Travel time per trial]{%
    \includegraphics[width=0.33\textwidth]{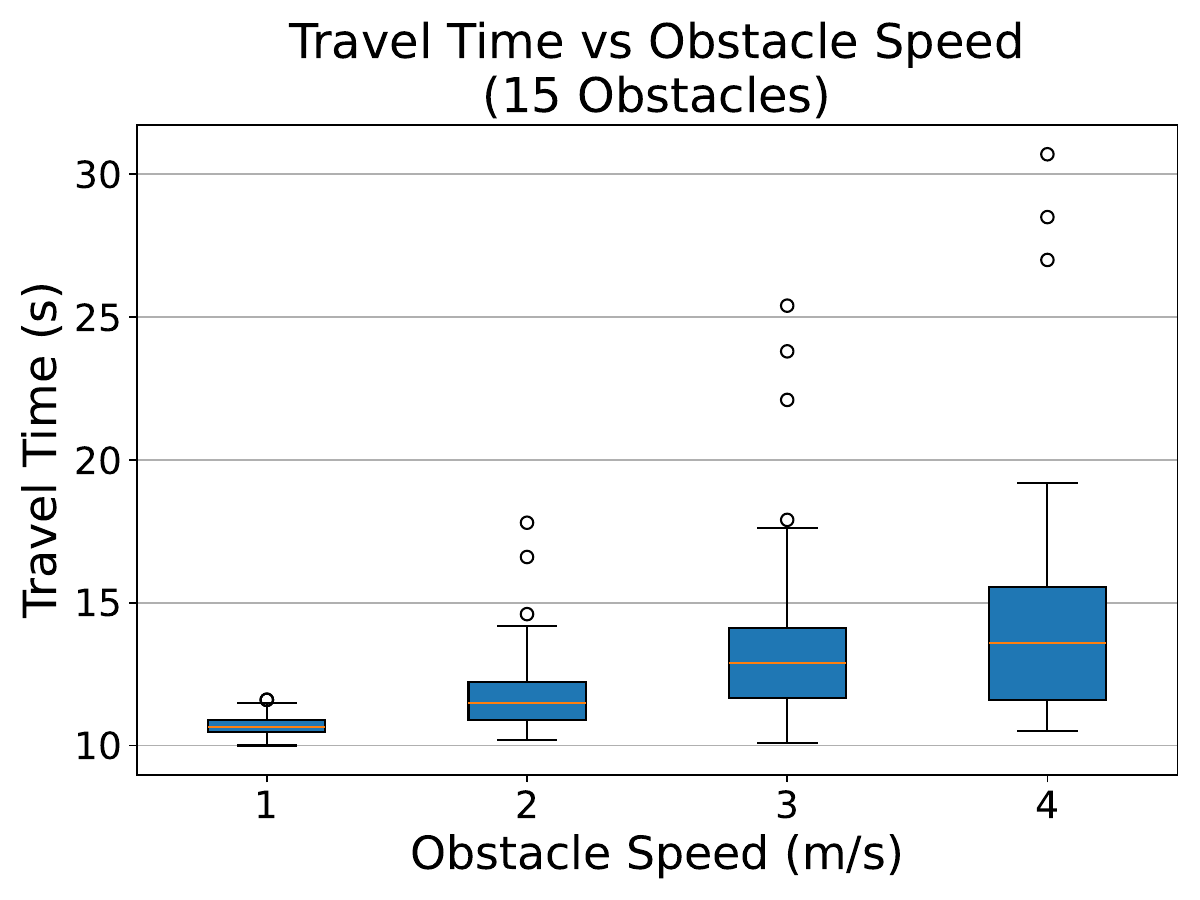}\label{fig:2D-c}%
  }\\ \vspace{10pt}
  \textbf{2D Case 1: Number of obstacles is fixed to be $15$ and  obstacle speed is varied from $1$m/s to $4$m/s.}
  \\[1ex]
  \subfloat[\hspace{0pt} Success rate over $100$ trials]{%
    \includegraphics[width=0.33\textwidth]{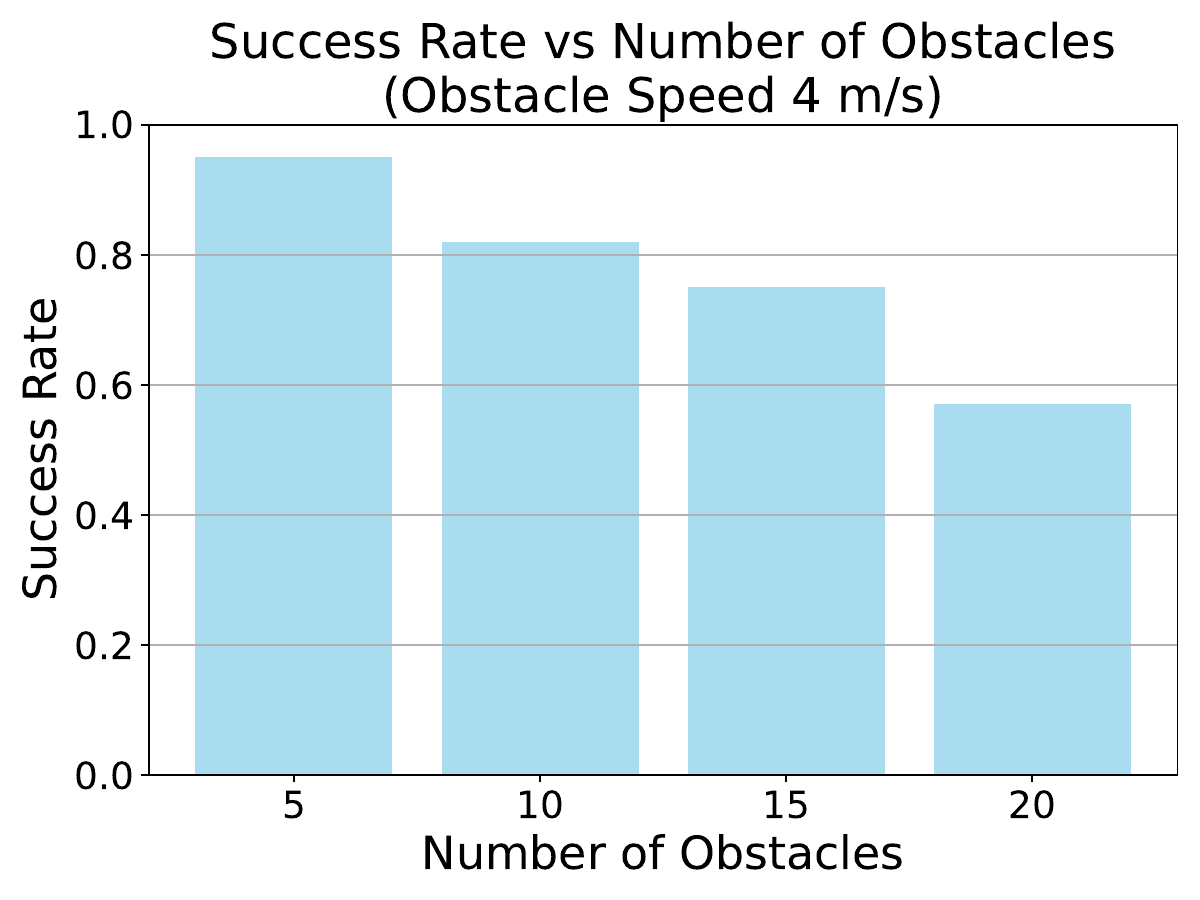}\label{fig:2D-d}%
  }\hfill
  \subfloat[\hspace{0pt} Average replanning time per trial]{%
    \includegraphics[width=0.33\textwidth]{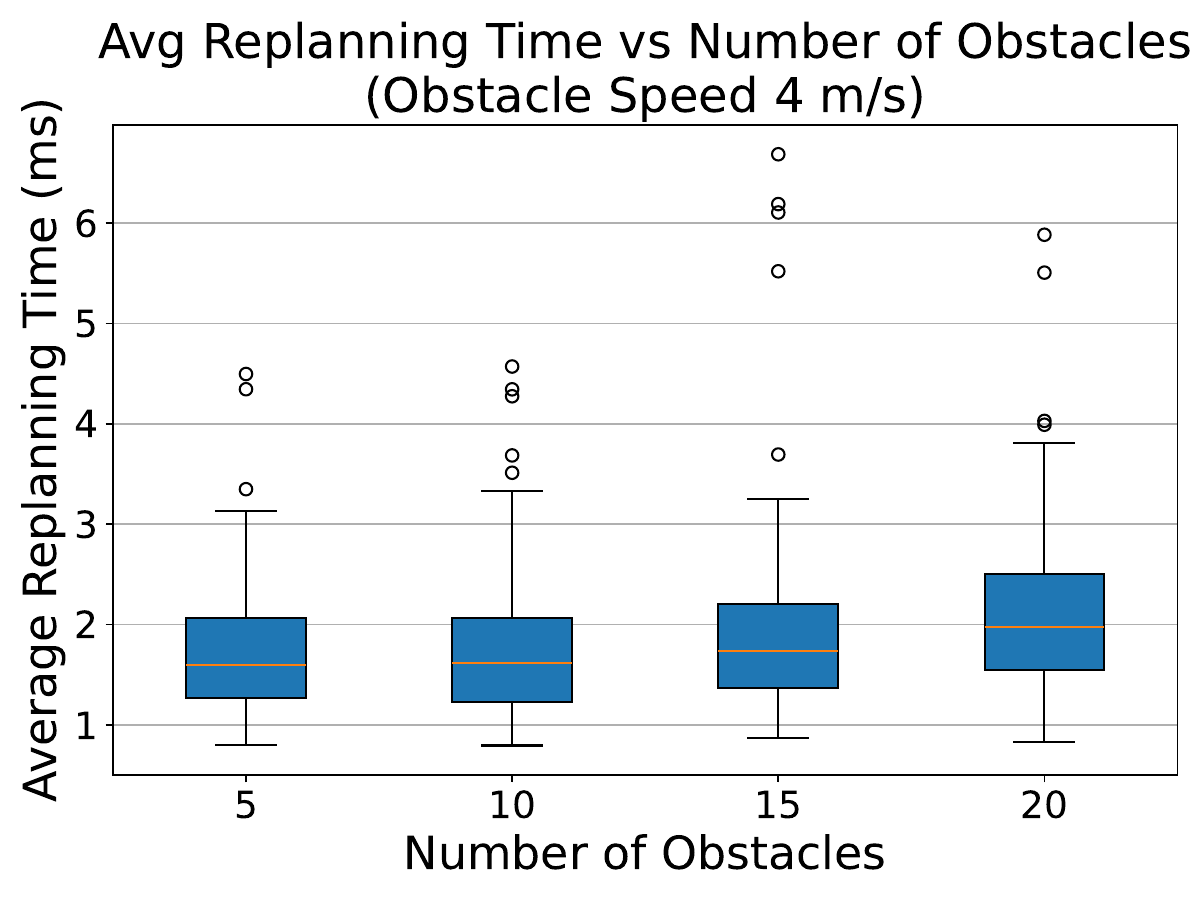}\label{fig:2D-e}%
  }\hfill
  \subfloat[\hspace{0pt}  Travel time per trial]{%
    \includegraphics[width=0.33\textwidth]{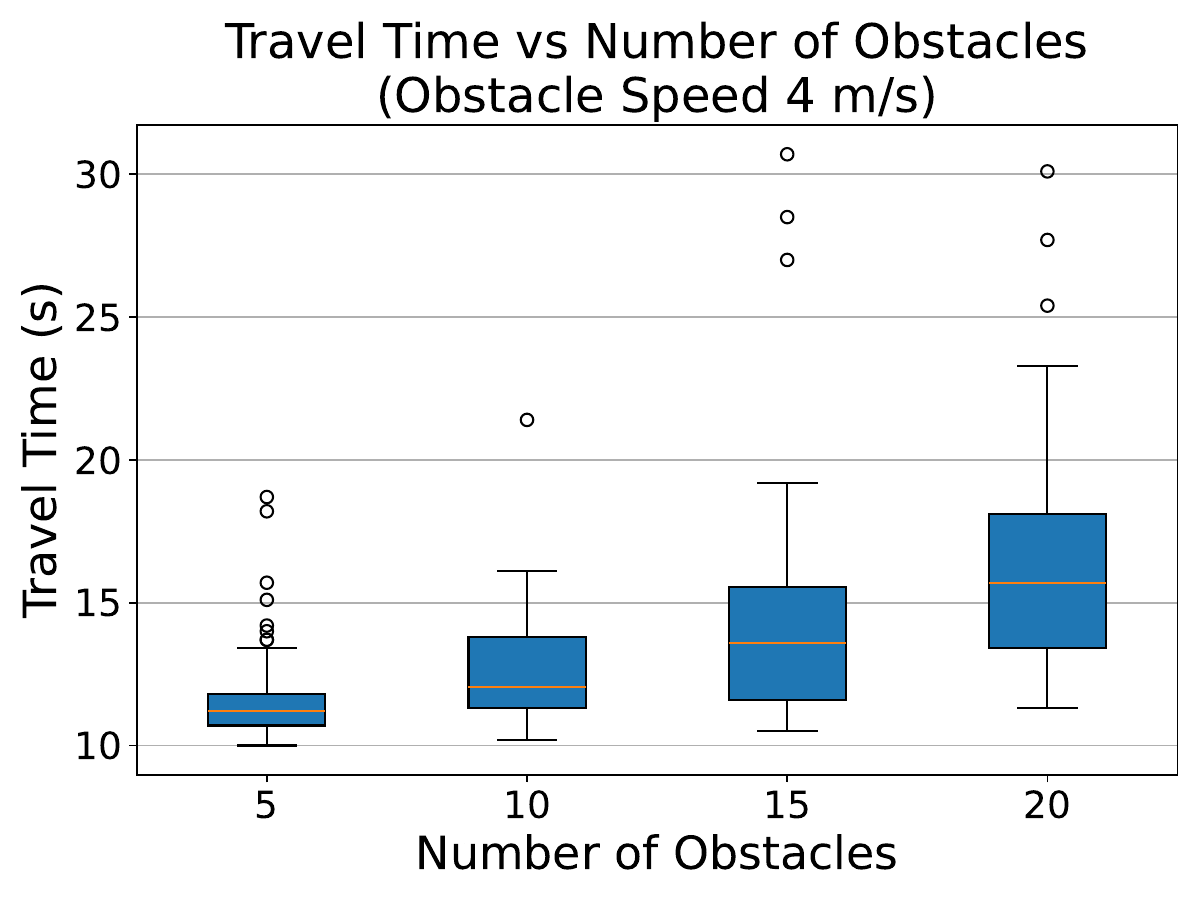}\label{fig:2D-f}%
  }\\ \vspace{10pt}
  \textbf{2D Case 2: Obstacle speed is fixed to be $4$m/s and number of obstacles is varied from $5$ to $20$.} \vspace{3pt}
  \caption{Performance results of simulation experiments for the two case studies of the 2D scenario.}
  \label{fig:2d-results}\vspace{-3pt}
\end{figure*}

\subsection{2D Scenario}
First, SMART-3D's performance is validated for a 2D scenario. Fig.~\ref{fig:2d-scenario-15-obs} shows an example of a $32$m $\times$ $32$m scenario with $15$ dynamic obstacles. In this scenario, the start and goal points of the robot are located  at ($2$m, $2$m) and ($30$m, $30$m), respectively. Each obstacle is considered to be circular with a radius of $0.5$m. The obstacles continuously move in a random heading from $[0,2\pi]$ for a random distance from $[0,10]$m, after which a new heading and distance are selected. The obstacles are allowed to overlap during traveling. 

Two case studies are conducted as follows. 

\vspace{3pt}
\subsubsection{2D Case 1: Number of obstacles fixed but speed varied} 
In the first 2D case study, the number of dynamic obstacles is fixed to be $15$, while the obstacle speed is varied. In any particular simulation trial, all obstacles move at the same constant speed chosen from the set $\{1,2,3,4\}$m/s. For each obstacle speed, we run $100$ trials, where in each trial the obstacles move randomly as described above. Then, a different speed is chosen, and the trials are repeated. In this manner, the statistical performance is recorded for four different speeds.  

\vspace{6pt}
\subsubsection{2D Case 2: Speed fixed but number of obstacles varied} 
In the second 2D case study, the speed of obstacles is fixed to be $4$m/s, while the number of dynamic obstacles is varied. In any particular simulation trial, the number of obstacles is chosen from the set $\{5,10,15,20\}$. For each number of obstacles, we run $100$ trials, where in each trial the obstacles move randomly as described above. Then, a different number of obstacles is chosen, and the trials are repeated. In this manner, the statistical performance is recorded for four different numbers of obstacles.

Fig. \ref{fig:2d-results} shows the performance results for the two case studies of the 2D scenario.  For case study $1$, Figs~\ref{fig:2D-a}, \ref{fig:2D-b} and \ref{fig:2D-c} show the plots of success rate over $100$ trials, average replanning time per trial and travel time per trial, respectively. The average replanning time is calculated over all replanning incidents in any trial. The results show that SMART-3D achieves similar performance as that of SMART~\cite{smart} for the 2D scenario. For obstacle speeds of $1$, $2$, $3$ and $4$m/s, SMART-3D achieves success rates of $1.0$, $1.0$, $0.88$ and $0.75$, respectively. 
The median of the average replanning time per trial falls between $1$ to $2$ milliseconds for every obstacle speed. The median travel times are approximately $10.7$s, $11.5$s, $12.9$s and $13.6$s, respectively. 
For case study $2$, Figs~\ref{fig:2D-d}, \ref{fig:2D-e} and \ref{fig:2D-f} show the plots of success rate over $100$ trials, average replanning time per trial, and travel time per trial, respectively. The success rate decreases as the number of obstacles is increased. The median of the average replanning time per trial falls between $1$ to $2$ milliseconds. The travel time increases with the number of obstacles due to more diversions and replannings. 

\begin{figure*}[!t]
  \centering
  \subfloat[\hspace{0pt} Success rate over $100$ trials]{%
    \includegraphics[width=0.33\textwidth]{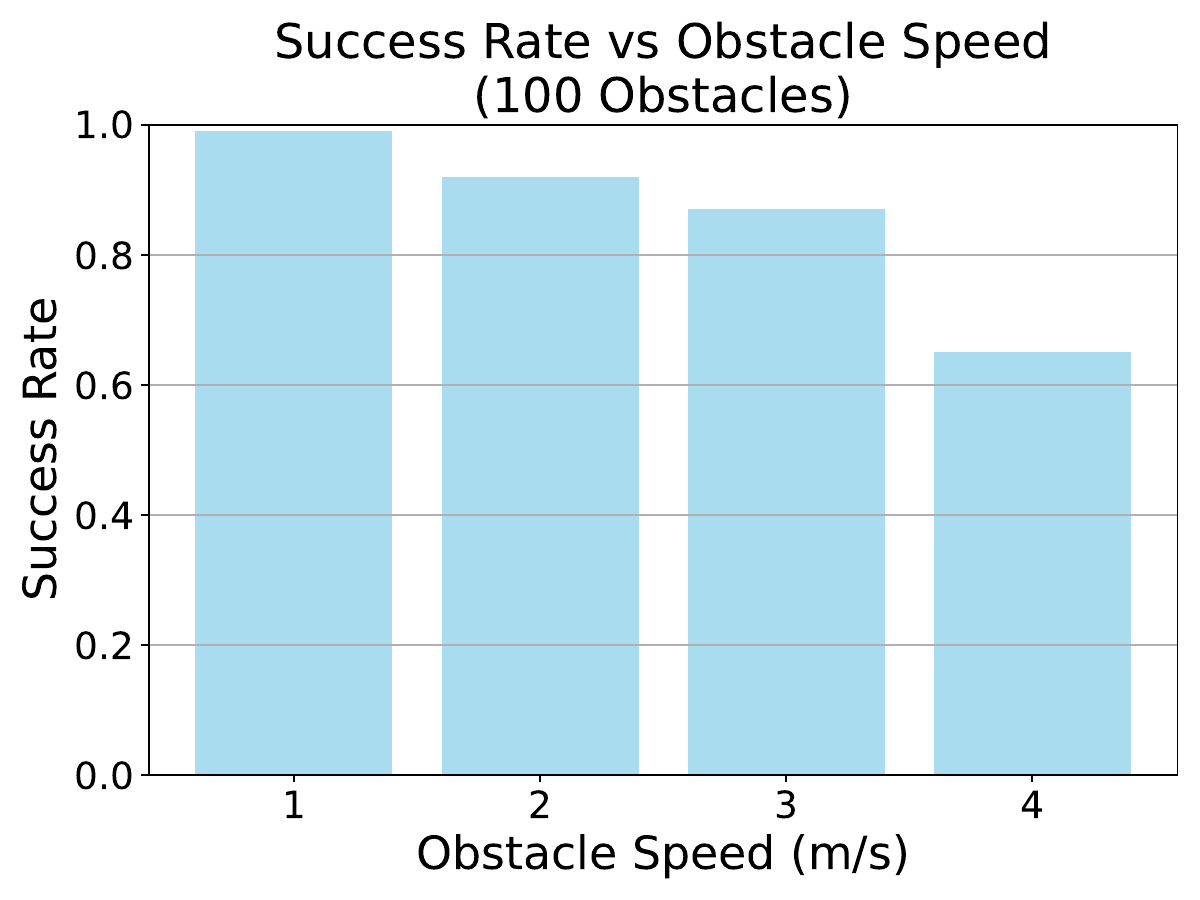}\label{fig:3D-a}%
  }\hfill
  \subfloat[\hspace{0pt} Average replanning time per trial]{%
    \includegraphics[width=0.33\textwidth]{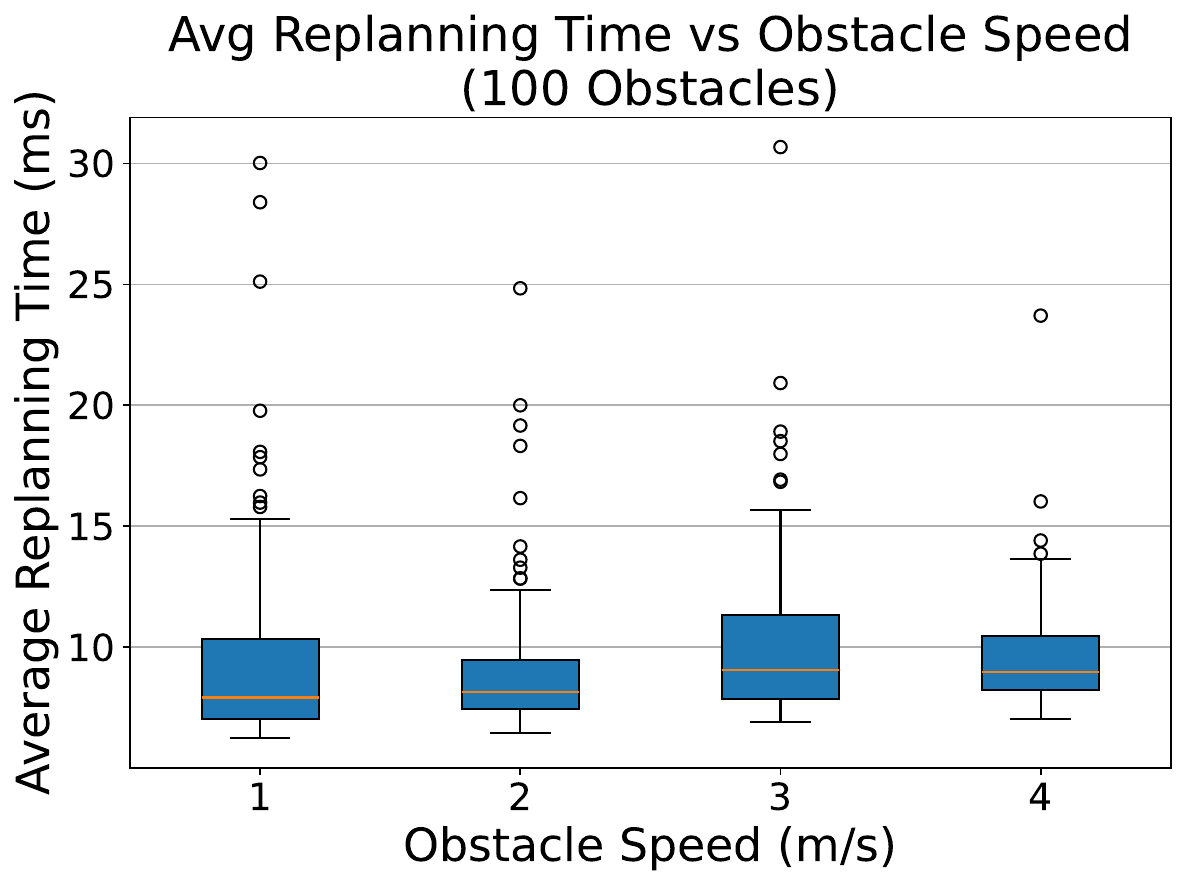}\label{fig:3D-b}%
  }\hfill
  \subfloat[\hspace{0pt} Travel time per trial]{%
    \includegraphics[width=0.33\textwidth]{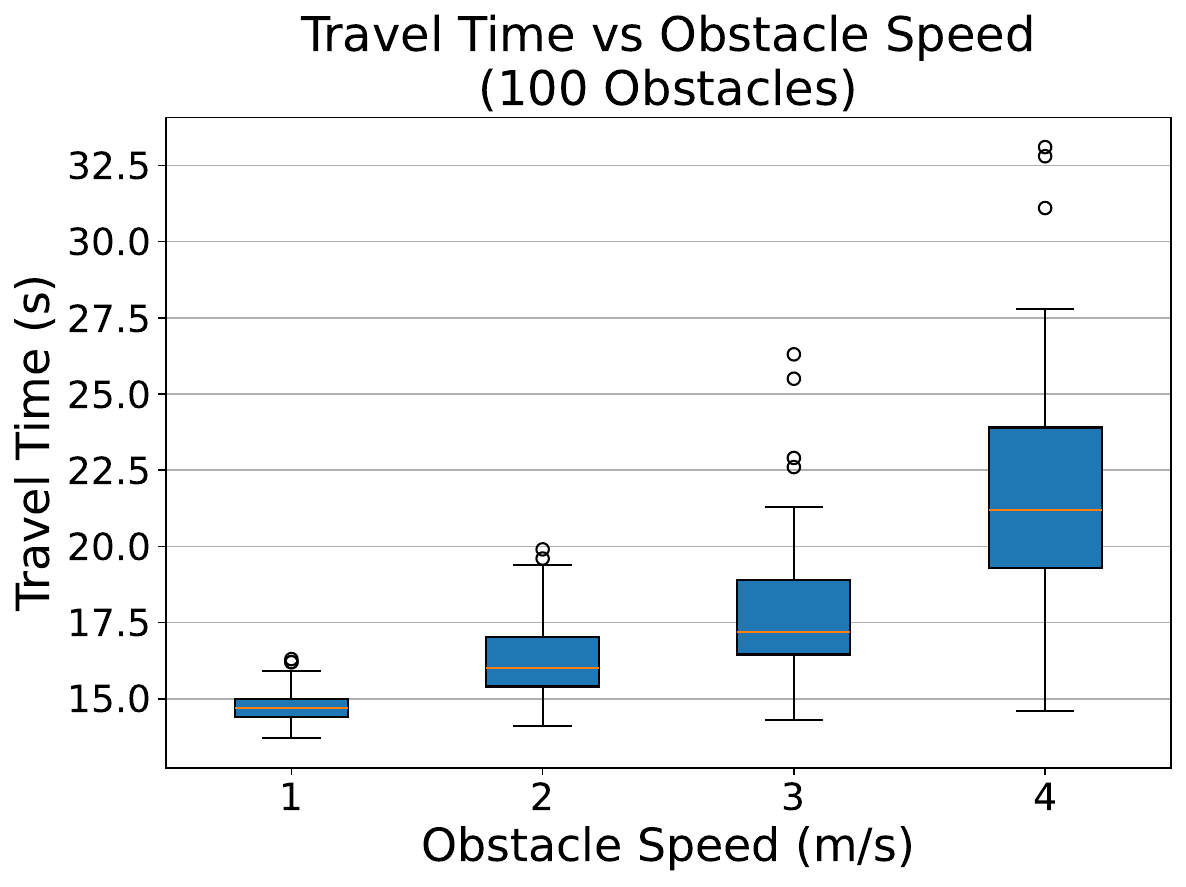}\label{fig:3D-c}%
  }\\ \vspace{8pt}
  \textbf{3D Case 1: Number of obstacles is fixed to be $100$ and  obstacle speed is varied from $1$m/s to $4$m/s.}
  \\[1ex]
  \subfloat[\hspace{0pt} Success rate over $100$ trials]{%
    \includegraphics[width=0.33\textwidth]{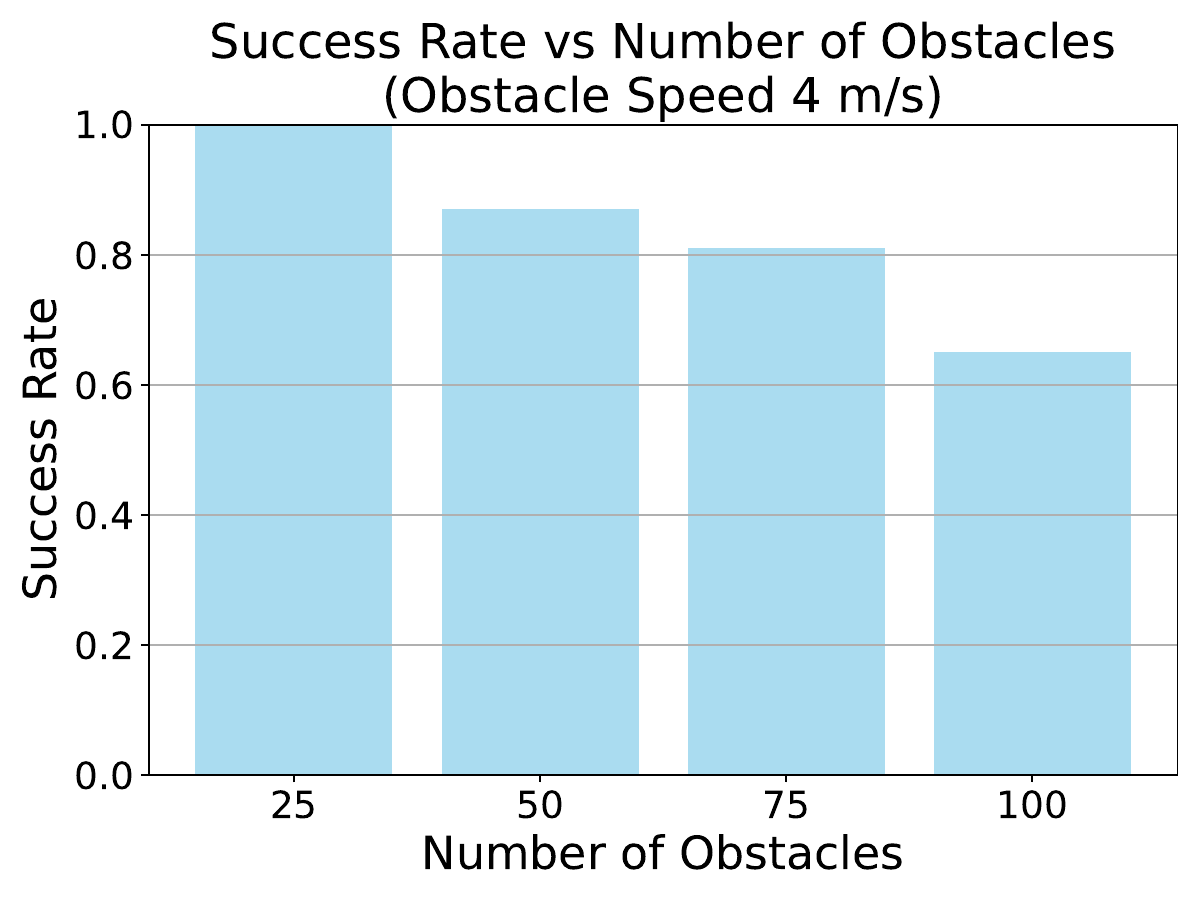}\label{fig:3D-d}%
  }\hfill
  \subfloat[\hspace{0pt} Average replanning time per trial]{%
    \includegraphics[width=0.33\textwidth]{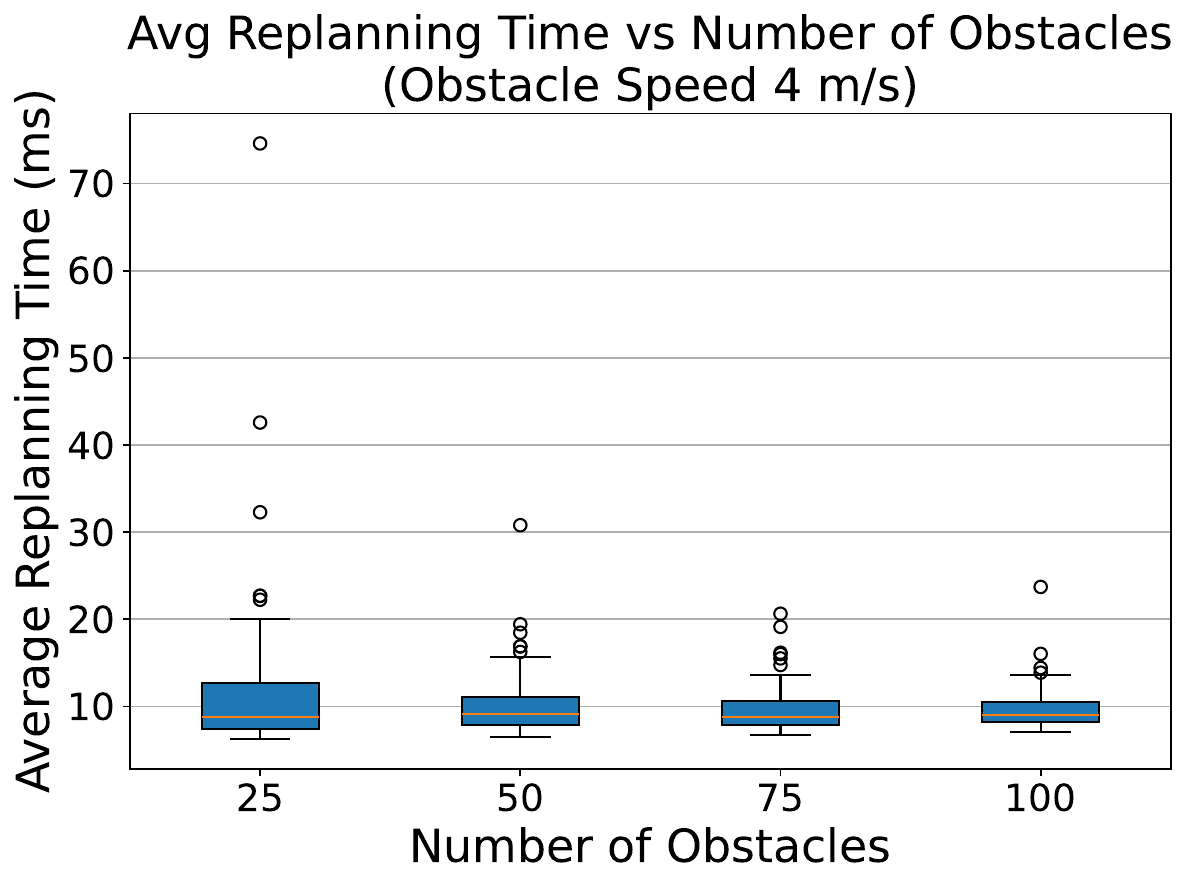}\label{fig:3D-e}%
  }\hfill
  \subfloat[\hspace{0pt} Travel time per trial]{%
    \includegraphics[width=0.33\textwidth]{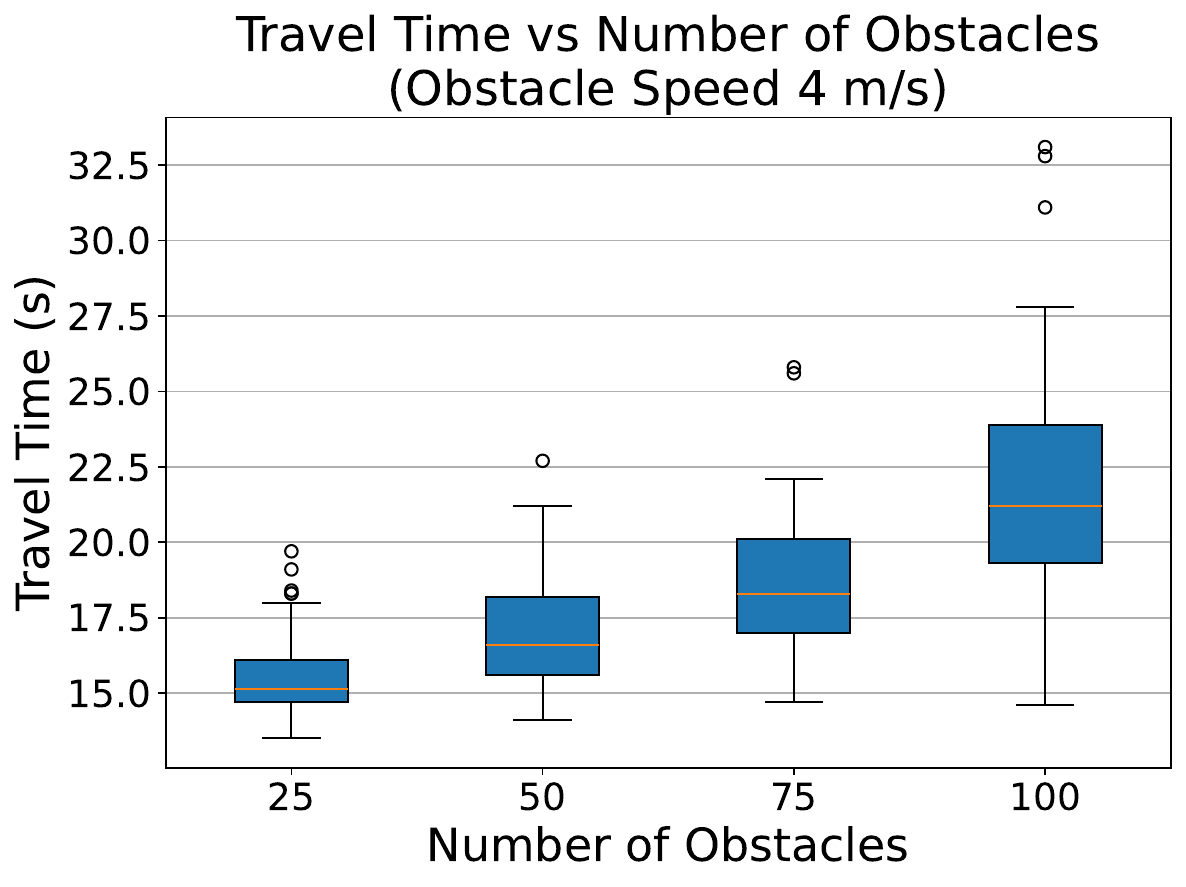}\label{fig:3D-f}
  }\\ \vspace{8pt}
  \textbf{3D Case 2: Obstacle speed is fixed to be $4$m/s and number of obstacles is varied from $25$ to $100$.} \vspace{3pt}
 \caption{Performance results of simulation experiments for the two case studies of the 3D scenario.}
  \label{fig:3d-results}\vspace{-3pt}
\end{figure*}

\vspace{-3pt}
\subsection{3D Scenario}
Next, SMART-3D's performance is validated for a 3D scenario. Fig.~\ref{fig:3d-scenario-100-obs} shows an example of a $32$m $\times$ $32$m $\times$ $32$m scenario with $100$ dynamic obstacles. In this scenario, the start and goal points of the robot are located at ($2$m, $2$m, $2$m) and ($30$m, $30$m, $30$m), respectively. Each obstacle is considered to be sphere with a radius of $0.5$m. The obstacles are initialized at random positions. They continuously move to randomly selected waypoints within the 3D space. The obstacles are allowed to overlap during traveling. 

Two case studies are conducted as follows. 

\vspace{6pt}
\subsubsection{3D Case 1: Number of obstacles fixed but speed varied}
In the first 3D case study, the number of dynamic obstacles is fixed to be $100$, while the obstacle speed is varied. In any particular simulation trial, all obstacles move at the same constant speed chosen from the set $\{1,2,3,4\}$m/s. For each obstacle speed, we run $100$ trials, where in each trial the obstacles move randomly as described above. Then, a different speed is chosen, and the trials are repeated. In this manner, the statistical performance is recorded for four different speeds. 

\vspace{6pt}
\subsubsection{3D Case 2: Speed fixed but number of obstacles varied} 
In the second 3D case study, the speed of obstacles is fixed to be $4$m/s, while the number of dynamic obstacles is varied. In any particular simulation trial, the number of obstacles is chosen from the set $\{25,50,75,100\}$. For each number of obstacles, we run $100$ trials, where in each trial the obstacles move randomly as described above. Then, a different number of obstacles is chosen, and the trials are repeated. In this manner, the statistical performance is recorded for four different numbers of obstacles. 

Fig.~\ref{fig:3d-results} shows the performance results for the two case studies of the 3D scenario. For case study $1$, Figs~\ref{fig:3D-a}, \ref{fig:3D-b} and \ref{fig:3D-c} show the plots of success rate over $100$ trials, average replanning time per trial and travel time per trial, respectively. Overall, SMART-3D achieves high success rates of around or above $90$\% for all obstacle speeds of $3$ m/s or less. As expected, the success rate decreases as the obstacle speed is increased. The median of the average replanning time per trial falls below $10$ milliseconds for every obstacle speed, which is slightly higher than the 2D scenario but fast enough for real-time applications. The median travel time increases with the obstacle speed due to more diversions and replannings. For case study $2$, Figs~\ref{fig:3D-d}, \ref{fig:3D-e} and \ref{fig:3D-f} show the plots of success rate over $100$ trials, average replanning time per trial and travel time per trial, respectively. They  show similar trends, where the success rate decreases with the number of obstacles, the median of the average replanning time stays below $10$ milliseconds, and the travel time increases with the number of obstacles.

\vspace{-3pt}
\section{Conclusion and Future Work}
This paper developed the SMART-3D algorithm which is an extension of the SMART algorithm~\cite{smart} to 3D spaces. By removing the need for a grid decomposition and replacing hot-spots with hot-nodes, SMART-3D scales well to a 3D environment with computational efficiency. The results for both 2D and 3D scenarios demonstrated that SMART-3D achieves high success rates and low replanning times, thus making it suitable for real-time applications. 

In future work, SMART-3D can be further extended to motion planning problems including time-risk optimization of non-holonomic robots~\cite{WG2025}, distributed multi-robot navigation~\cite{song2020care}, and object manipulation using robotics arms with complex configuration spaces.

\vspace{-3pt}
\appendix
\subsection*{Code Availability}
The code for SMART-3D is available at:  
\url{https://github.com/LinksLabUConn/SMART3D}.

\vspace{-6pt}
\bibliographystyle{IEEEtran}   
\bibliography{smart3d}

\end{document}